\documentclass[acmlarge]{acmart}

\AtBeginDocument{%
  }

\setcopyright{acmlicensed}
\copyrightyear{2024}
\acmYear{2024}
\acmDOI{XXXXXXX.XXXXXXX}

\acmJournal{JACM}
\usepackage{multirow}
\usepackage{amsfonts}
\usepackage{multirow}       
\usepackage{comment}
\usepackage{balance}
\usepackage{csquotes}
\usepackage{comment,adjustbox}
\usepackage{xcolor, soul}
\sethlcolor{yellow}
\usepackage{flushend}
\begin{document}

%%
%% The "title" command has an optional parameter,
%% allowing the author to define a "short title" to be used in page headers.
\title{Social Media Authentication and Combating Deepfakes using Semi-fragile Invisible Image Watermarking}

%%
%% The "author" command and its associated commands are used to define
%% the authors and their affiliations.
%% Of note is the shared affiliation of the first two authors, and the
%% "authornote" and "authornotemark" commands
%% used to denote shared contribution to the research.
\author{Aakash Varma Nadimpalli}
\orcid{0009-0003-7254-3207}
\affiliation{%
  \institution{Wichita State University}
 % \streetaddress{1 Th{\o}rv{\"a}ld Circle}
  \city{Wichita, Kansas}
  \country{USA}}
\email{axnadimpalli@shockers.wichita.edu}

\author{Ajita Rattani}
\orcid{0000-0002-1541-8202}
\affiliation{%
  \institution{University of North Texas}
  \city{Denton, Texas}
  \country{USA}}
  \email{ajita.rattani@unt.edu}

%%
%% By default, the full list of authors will be used in the page
%% headers. Often, this list is too long, and will overlap
%% other information printed in the page headers. This command allows
%% the author to define a more concise list
%% of authors' names for this purpose.
\renewcommand{\shortauthors}{Nadimpalli and Rattani}

%%
%% The abstract is a short summary of the work to be presented in the
%% article.
\begin{abstract}
%The growing issue of deepfakes and manipulated media is exacerbated by the advancement of complex image and video synthesis technologies. 

With the significant advances in deep generative models for image and video synthesis, Deepfakes and manipulated media have raised severe societal concerns. Conventional machine learning classifiers for deepfake detection often fail to cope with evolving deepfake generation technology and are susceptible to adversarial attacks. 
Alternatively, invisible image watermarking is being researched as a proactive defense technique that allows media authentication by verifying an invisible secret message embedded in the image pixels. A handful of invisible image watermarking techniques introduced for media authentication have proven vulnerable to basic image processing operations and watermark removal attacks. 
In response, we have proposed a semi-fragile image watermarking technique that
embeds an invisible secret message into real images for media authentication.
Our proposed watermarking framework is designed to be fragile to facial manipulations or tampering while being robust to benign
image-processing operations and watermark removal attacks. This is facilitated through a unique architecture of our proposed technique consisting of critic and adversarial networks that enforce high image quality and resiliency to watermark removal efforts, respectively, along with the backbone encoder-decoder and the discriminator networks. 
This allows images shared over the Internet to retain the verifiable watermark as long as facial manipulations
or any other Deepfake modification technique is not applied. 
Thorough experimental investigations on SOTA facial Deepfake datasets demonstrate that our proposed model can embed a $64$-bit secret as an imperceptible image watermark that can be recovered with a high-bit recovery
accuracy when benign image processing operations are applied while being non-recoverable when unseen Deepfake manipulations are applied.
In addition, our proposed watermarking technique demonstrates high resilience to several white-box and black-box watermark removal attacks. Thus, obtaining state-of-the-art performance.

\end{abstract}

%%
%% The code below is generated by the tool at http://dl.acm.org/ccs.cfm.
%% Please copy and paste the code instead of the example below.
%%
\begin{CCSXML}
<ccs2012>
   <concept>
       <concept_id>10010147.10010371.10010382</concept_id>
       <concept_desc>Computing methodologies~Image manipulation</concept_desc>
       <concept_significance>500</concept_significance>
       </concept>
   <concept>
       <concept_id>10010147.10010178.10010224</concept_id>
       <concept_desc>Computing methodologies~Computer vision</concept_desc>
       <concept_significance>300</concept_significance>
       </concept>
   <concept>
       <concept_id>10002978.10003029.10003032</concept_id>
       <concept_desc>Security and privacy~Social aspects of security and privacy</concept_desc>
       <concept_significance>500</concept_significance>
       </concept>
 </ccs2012>
\end{CCSXML}
\ccsdesc[500]{Computing methodologies~Image manipulation}
\ccsdesc[300]{Computing methodologies~Computer vision}
\ccsdesc[500]{Security and privacy~Social aspects of security and privacy}

%%
%% Keywords. The author(s) should pick words that accurately describe
%% the work being presented. Separate the keywords with commas.
\keywords{Facial Manipulations, Deepfakes, Media Authentication, Watermarking.}

%\received{20 February 2007}
%\received[revised]{12 March 2009}
%\received[accepted]{5 June 2009}

%%
%% This command processes the author and affiliation and title
%% information and builds the first part of the formatted document.
\maketitle

\section{Introduction}
\label{introduction}
Media authentication refers to the process of verifying the authenticity and integrity of digital media such as images, videos, audio recordings, or text documents~\cite{Fridrich_2009}. With the advancement in generative models, combined with the widespread availability of vast datasets, there is a rise in digital manipulation tools and techniques that have enabled the creation of high-quality and convincing AI-generated synthetic media (such as face, audio, and text) known as \textit{Deepfakes}~\cite{Tolosana2020deepfakesAB,NGUYEN2022103525,1282}. Apart from many creative and artistic uses of deepfakes~\cite{Chan2018EverybodyDN}, many harmful uses range from non-consensual pornography to disinformation campaigns intended to sow civil unrest and disrupt democratic elections. Deepfakes have been flagged as a top AI threat to society~\cite{NGUYEN2022103525, 9157215}. %Apart from many creative and artistic uses of deepfakes~\cite{Chan2018EverybodyDN}, many harmful uses range from non-consensual pornography to disinformation campaigns meant to sow civil unrest and disrupt democratic elections. deepfakes have been flagged as a top AI threat to society~\cite{NGUYEN2022103525, 9157215}.

In this context, several deepfake generation techniques based on facial manipulation (forgery) have been proposed~\cite{NGUYEN2022103525,1282}. These facial manipulations or forgery techniques depict human subjects with altered identities (identity swap), attributes, or malicious actions and expressions (face reenactment) in a given image or video. Specifically, identity or face swapping is the task of transferring a face from the source to the target image~\cite{Nirkin2019FSGANSA}. Attribute manipulation~\cite{He2017AttGANFA,Choi2017StarGANUG} is a fine-grained facial manipulation obtained by modifying simple attributes (e.g., hair color, skin tone, and gender). Similarly to identity swap, face reenactment~\cite{Nirkin2019FSGANSA} involves a facial expression swap between source and target facial images. 
These facial manipulation tools are easily abused by malicious users, \emph{with little to no technical knowledge}, to manipulate facial images of the user, resulting in a threat to privacy, reputation, and security. In fact, several smartphone-based applications have such attribute modifications in the form of filters. For example, FaceApp\footnote{\url{https://www.faceapp.com/}}, a popular smartphone application, modifies an uploaded image based on the selected attribute that can be edited using a slider to regulate the magnitude of the change. The entire process of facial modification can be easily accomplished within five minutes using these applications or other pretrained models available in the online repositories.

Consequently, every year the \textbf{volume} of facial Deepfakes on social media has witnessed a significant rise. For instance, in $2023$ alone, about $500,000$ deepfake videos were added to social media, marking a substantial rise from previous years. In $2021$, there were approximately $14,678$ deepfake videos online, which itself was double the number from $2018$~\cite{Zaidi_2023}.

With this staggering growth of facial manipulation-based deepfake content in social media, it has become increasingly important to ensure the media's authenticity against malicious tampering. The classical \textit{forensic approach} for media authentication against facial manipulation includes running \textbf{an automated deepfake detector}~\cite{8630787,dolhansky2020deepfake,nadimpalli2022improving}. Common Deepfake detectors include pre-trained machine learning-based binary baselines that aim to distinguish between real and deepfake data based on visual artifacts, blending boundaries, attention module, and motion analysis~\cite{He2016DeepRL,8099678,7780677,9157215,9578910,9577592}. %Apart from CNN baselines, other deepfake detection techniques include detecting blending boundaries~\cite{9157215}, lip-syncing~\cite{9578910}, and multi-attentional models~\cite{9577592}. %These aforementioned 
%\emph{passive} detection techniques are an \emph{ex-post forensics countermeasure} and are still in their early stage~\cite{Wang2022AntiForgeryTA,Wang_2022_CVPR}. They suffer from poor detection accuracy~\cite{Peng2022DFGC2T,8099678}, 
%cross-dataset generalizability~\cite{9577592, Nadimpalli2022OnIC}, and obtain biased performance across demographic attributes such as gender and race~\cite{Nadimpalli2022GBDFGB, Trinh2021AnEO}.
%aim to  distinguish between Real and DeepFake data. The popular DeepFake detection techniques include training convolution neural network (CNN) based binary classification baselines~\cite{He2016DeepRL,8099678,7780677}, detecting blending boundaries~\cite{9157215}, lip-syncing~\cite{9578910}, and multi-attentional model~\cite{9577592}. 
These \emph{passive} Deepfake detection techniques, an \emph{ex-post forensics countermeasure}, and are still in their early stage~\cite{Wang2022AntiForgeryTA,Wang_2022_CVPR} as these techniques suffer from poor detection accuracy~\cite{Peng2022DFGC2T,8099678}, 
cross-dataset generalizability~\cite{9577592,nadimpalli2022improving}, obtain differential performance across demographic attributes such as gender and race~\cite{nadimpalli2022gbdf, Trinh2021AnEO}, and are vulnerable to adversarial attacks~\cite{nadimpalli2022gbdf, Trinh2021AnEO}. Further, they fail to cope with ever-evolving deepfake generation techniques.
%Further, these passive techniques cannot completely prevent the negative impact as \textit{debunking}\footnote{https://blogs.microsoft.com/on-the-issues/2020/09/01/disinformation-deepfakes-newsguard-video-authenticator/} deepfakes from social media takes time. Therefore, they fail in blocking the disinformation spread in advance and harm to the reputation of the victim is ever-lasting. 

\begin{figure*}[tbp]
\centerline{\includegraphics[width=1.0\textwidth]{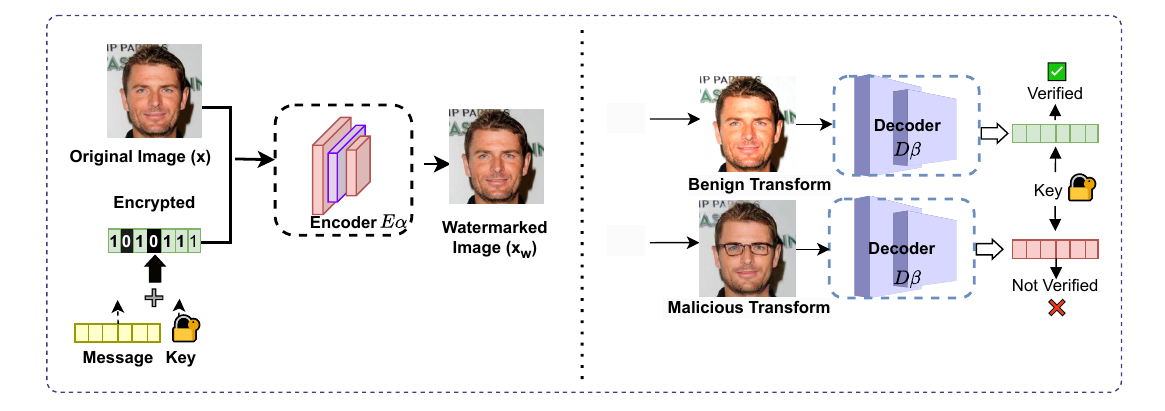}}
\caption{Overview of our proposed framework that involves embedding a secret encrypted message into an image using an encoder-decoder style network for the purpose of media authentication. This watermark is imperceptible to the human eye and resistant to typical image alterations and watermark removal attacks, but it is vulnerable to malicious facial transformations i.e., Deepfakes.}
\label{fig1}
\end{figure*}

Alternatively, \textbf{watermarking} is being actively researched as a \textit{proactive} defense technique because it involves embedding invisible markers or signatures into authentic media content such as images or videos. These markers are unique to the creator and can help in identifying the authenticity of the content~\cite{1286417,tancik2020stegastamp,zhu2018hidden} by matching the watermark message retrieved from the media to the original embedded watermarked message. 
Thus, by watermarking content, before it gets shared or distributed, creators can take preventive measures against malicious use or alteration of their work (as indicated by alteration to the watermark retrieved from the media). Hence, promoting and regulating the \textit{responsible and ethical use} of AI, an important initiative of government leaders and the legislature~\cite{zhao2023generative}. \textit{Invisible watermarks} are preferred because they preserve the image quality and it is less likely
for a layperson to tamper with it.
%Watermarking media content involves embedding a piece of information often called a "watermark" into digital media such as images, videos, audio, or documents. This watermark can be visible, like a logo or text overlay on the media, or invisible and encoded in a way that it can be detected/ retrieved via specialized software. %The watermarking could be used to protect copyright, prove ownership, authenticate content, track distribution, and deter unauthorized use or reproduction of the media.  
Traditional image watermarking~\cite{info11020110} techniques typically transform domain coefficients of an image, using various transforms such as the Discrete Cosine Transform (DCT) and Discrete Fourier Transform (DFT) for watermark embedding. 
%However, these conventional watermarking techniques are not resilient to transformations and removal attacks~\cite{info11020110} as compared to deep-learning-based watermarking approaches~\cite{zhong2023deep}.
%Additionally, media images that have been watermarked using traditional methods also tend to obtain higher error rates on state-of-the-art DeepFake detectors. This is attributed to alterations in the high-frequency components of images, which are crucial for identifying the manipulation signatures that most DeepFake detectors rely on~\cite{app132111852}.
 Deep-learning-based techniques such as StegaStamp~\cite{tancik2020stegastamp}, and HiDDeN~\cite{zhu2018hidden} have emerged as an efficient solution over traditional image watermarking in terms of an end-to-end solution for efficient message embedding~\cite{luo2020distortion,tancik2020stegastamp,zhu2018hidden,NIPS2017_838e8afb,hayes2017generating}. 
 %These approaches substitute image transform-based watermark hiding techniques with neural network-based encoding techniques %~\cite{luo2020distortion,tancik2020stegastamp,zhu2018hidden,NIPS2017_838e8afb,hayes2017generating}. 
 However, these aforementioned watermarking techniques are also either fragile (i.e., watermark message is altered) to basic image processing operations such as compression and color adjustments~\cite{zhu2018hidden,1286417} or overly robust to malicious transform which deters manipulated media detection~\cite{tancik2020stegastamp}. A study in~\cite{10.1145/3625547} proposed GAN-based visible watermarking for media authentication. However, visible watermarks are more likely for the layperson to tamper with.

 %the point that the secret can be recovered even after occluding major portions of the embedded image~\cite{tancik2020stegastamp}.

Importantly, for efficient media authentication and detection of manipulations, a \textbf{semi-fragile} invisible watermarking scheme is required that is robust to benign image transformations (such as contrast enhancement) and vulnerable to malicious transformations such as face-swapping-based Deepfakes. In other words, watermark content/messages retrieved from the media remain unaltered to the benign and altered to the malicious transforms (facial manipulations).
Resiliency to benign transforms ensures that the authenticity of the digital media can be validated in the presence of basic image processing operations such as compression, resizing, and color adjustment, which are often applied during image sharing, editing, or storing. At the same time, the vulnerability of the embedded watermark to malicious transformations is required to detect potential forgery or unauthorized modifications.
%Thus robustness to these benign transformations allows the watermark to remain intact and verifiable unless the image undergoes intentional, malicious tampering, at which point the watermark should degrade to indicate potential forgery or unauthorized modifications. 
%Thus, there is a need for \textit{semi-fragile} watermarking technique for social media authentication which is robust to benign and vulnerable to malicious transformations. 
Traditional semi-fragile watermarking techniques~\cite{zhu2018hidden,tancik2020stegastamp,1286417} that function within the transform domain, such as semi-fragile DCT~\cite{1286417}, are vulnerable to high-level semantic transformations to images/media and may struggle to keep up with new or advanced manipulation techniques as technology advances. Additionally, the effectiveness of these techniques largely relies on the specific transform domain selected and the parameters set for embedding the watermark.
%This balance helps in distinguishing between legitimate and tampered content effectively.
A recent work~\cite{neekhara2022facesigns} proposes a first-of-its-kind deep-learning-based semi-fragile invisible watermarking scheme called \emph{FaceSigns}, that is based on encoder-decoder style model and can withstand benign transformations but is vulnerable to malicious deepfake transformations for manipulated media detection. Although invisible watermarks are less likely for a layperson to tamper with, abusers are not laypersons. They will make a deliberate attempt to remove these watermarks. Therefore, the injected
invisible watermark must be robust to evasion (watermark removal) attacks.
However, there is a \textit{notable gap} in this research regarding the model's susceptibility to watermark removal attacks. Furthermore, there is also limited investigation into the model's ability to generalize to unseen facial manipulations obtained using different generative techniques.

This paper \textbf{aims} to introduce a novel semi-fragile invisible watermarking scheme for social media authentication that can generate high-quality watermarked images.
At the same time, demonstrate resilience (i.e., the retrieved watermark remains intact) to both known and unknown benign transformations. In addition, the watermarking is vulnerable to unknown malicious facial transformations. In addition, the embedded watermark using our proposed model is robust against watermark removal attacks, including white-box attacks~\cite{goodfellow2014explaining,carlini2017towards,athalye2018obfuscated} and black-box attacks~\cite{zhao2023generative}. Thus, addressing the limitation of the existing semi-fragile watermarking technique. This is facilitated through a unique architecture of our proposed model consisting of critic and adversarial networks together with their corresponding novel loss functions, along with the backbone encoder-decoder and the discriminator network. Semi-fragile watermarking was chosen for this authentication method because it effectively balances robustness and sensitivity, making it ideal for deepfake detection. In other words, this watermarking scheme withstands benign transformations, such as resizing or compression, when applied to the watermarked images, without triggering false positives, while remaining sensitive enough to detect malicious alterations such as deepfakes. This ensures reliable and accurate detection of deepfakes, which is crucial to preserving the integrity of social media images. Figure~\ref{fig1} illustrates the overview of the proposed approach in embedding semi-fragile invisible encrypted watermarks in facial images that withstand benign transformations and are vulnerable to malicious transformations, for social media authentication. 
%To this aim, the contributions of this work are as follows:%Additionally, our aim is to ensure the generalizability of the proposed method to unseen manipulation techniques and to enhance its robustness against various threat models.
The technical \textbf{contributions} of our work are as follows:
\begin{enumerate}
\item A novel semi-fragile invisible facial image watermarking technique for social media authentication and for combating deepfakes, that offers superior imperceptibility and is resilient to adversarial watermark removal attacks.

\item Evaluation of the model's imperceptibility over other SOTA watermarking models in terms of peak signal-to-noise ratio~(PSNR)  and structural similarity index metrics~(SSIM).
%\item Extensive experiments are done on both robustness and fragility by using various unseen benign and malicious image transformations (not used during training).
\item Robustness analysis of the proposed model against unknown benign and malicious facial manipulation using different generative models.

\item Robustness analysis of the proposed model against watermark removal attacks using various white-box based (such as Fast Gradient Sign Method (FGSM)~\cite{goodfellow2014explaining}, Carlini\& Wagner~\cite{carlini2017towards}, Backward Pass Differentiable Approximation~(BPDA)~and Expectation Over Transformation~(EOT)~\cite{athalye2018obfuscated}) and black-box based watermark removal attacks (based on VAE Embedding and Reconstruction~\cite{zhao2023generative}). % please refer section~\ref{Robustness and threat analysis} for more detailed information.

\item Through evaluation on the SOTA facial image datasets, namely, FaceForensics++~\cite{9010912}, CelebA~\cite{liu2015faceattributes} and IMDB-WIKI~\cite{7406390}, widely adopted for facial manipulation-based deepfake generation and detection.

\item Ablation study to better understand the impact of each module (network) used in our proposed model and threat model for the adversarial attacks against our proposed model.
%To this
%front, we separately added Critic and Adversary modules to
%the U-Net (Baseline), please refer section~\ref{Ablation} for more detailed information.
\end{enumerate}

\noindent The \textit{pros and cons} of our proposed work are as follows: Our work presents a novel semi-fragile invisible watermarking scheme for social media authentication, generating high-quality watermarked images that withstand both known and unknown benign transformations while remaining vulnerable to malicious facial manipulations. This is facilitated through our proposed model's unique architecture, which combines critic and adversarial networks with novel loss functions, a backbone encoder-decoder, and a discriminator network. This innovative design enables our watermarking scheme to overcome the shortcomings of previous watermarking methods, including susceptibility to watermark removal attacks, white-box attacks, and limited generalizability to unseen facial manipulations. Our work has two primary limitations. Firstly, the complexity of our model necessitates advanced hardware and GPU support, which we plan to address in future iterations by optimizing the model for practical deployment. Secondly, we were unable to simulate all the potential attacks described in the threat model in Section~\ref{Threat Model}, but we intend to expand our model to better withstand these threats in future work.

This paper is organized as follows. Section~\ref{Related work} discusses the prior work on facial manipulation generation and passive deepfake detection, and image watermarking for media authentication. Section~\ref{Proposed Method} discusses our proposed methodology of semi-fragile invisible watermarking technique. Section~\ref{Experimental validations} discusses the implementation and experimental details including the datasets used and the performance evaluation metrics. Section~\ref{Imperceptibility and Capacity} discusses the imperceptibility and capability measures of various watermarking schemes. Section~\ref{Robustness and Fidelity} discusses the comparative evaluation of the robustness and fidelity of different watermarking methods by exposing them to unseen benign and malicious transformations and calculating the retrieved Bit Recovery Accuracy~(BRA). Section~\ref{Robustness and threat analysis} discusses the adversarial attacks and threat analysis of the proposed models against different white-box and black-box adversarial attacks along with the threat model. Section~\ref{Ablation} discusses the ablation study to better understand the impact of each module used in our proposed model. Section~\ref{conclusion} discusses the conclusion and future research directions.

\section{Related Work}
\label{Related work}

\subsection{Facial Manipulation Generation and Passive Deepfake Detection}
\label{Passive Deepfake Detection}

Facial manipulations are categorized primarily into three groups: identity swap~\cite{Nirkin2019FSGANSA}, attribute manipulation~\cite{Choi2017StarGANUG,He2017AttGANFA}, and expression swap~\cite{thies2016face2face,thies2019deferred}. 
Generative models such as auto encoders~\cite{bank2023autoencoders} (such as Faceswap), Generative Adversarial Networks (GANs)~\cite{goodfellow2014generative} (as FSGAN~\cite{Nirkin2019FSGANSA} and AttGAN~\cite{He2017AttGANFA}), and Diffusion Models (stable diffusion)~\cite{ho2020denoising} are used to create highly realistic fake content, including non-existent faces or altering existing ones~\cite{podell2024sdxl}. Among all, GAN is the most commonly adopted model for the generation of facial manipulation because it excels at generating high-quality realistic images that mimic the distribution of the original dataset. %This popularity stems from their unique structure, where two neural networks—the generator and the discriminator—compete against each other. %while diffusion-based models are at their initial stages.

%In nutshell, auto-encoders are used for FaceSwap creation by employing a two-part neural network structure: an encoder and a decoder. The encoder compresses an input face image into a compact latent representation, capturing the essential features while reducing data dimensionality. The decoder then reconstructs the face image from this latent space, but it can be trained to reconstruct the facial features of a different person, effectively swapping faces. Similarly, GANs (Generative Adversarial Networks) and diffusion models are used for face-swapping by leveraging their ability to generate and modify images. GANs consist of two parts: a generator that creates images and a discriminator that evaluates them. For face-swapping, the generator learns to output an image with the facial features of one person superimposed onto another person's face, trained against a discriminator that tries to detect fakes. Diffusion models iteratively add and remove noise to transform one face into another, refining the output through a process that gradually shifts from one identity to another, ensuring realistic transformation.

%These manipulations can significantly impact various sectors, from entertainment to spreading misinformation on social networks. The rapid evolution of deep learning has made it easier to synthesize or manipulate faces, raising public concern and prompting research into detection methods.

The majority of methods currently used for DeepFake detection are based on Convolutional Neural Networks (CNNs) (such as VGG16, ResNet50, ResNet101, ResNet152, and Xception) based binary baselines~\cite{Tolosana2020deepfakesAB, Li_2019_CVPR_Workshops,Nadimpalli2023FacialFD,nadimpalli2022improving,nadimpalli2022gbdf}. %These include using CNN architectures~\cite{Li_2019_CVPR_Workshops} like VGG16, ResNet50, ResNet101, ResNet152, and Xception for identifying manipulations in facial regions and their surroundings. 
Other approaches include Long-Short-term Memory (LSTM) networks~\cite{CHEN202258} to analyze spatio-temporal data, using facial and behavioral biometrics~\cite{Dong2020IdentityDrivenDD,Agarwal2019ProtectingWL,9717407}, examining inconsistencies in mouth movements~\cite{9578910}, multi-attentional~\cite{9577592} models that focus on different parts of the image, and $F^{3}$ -Net~\cite{10.1007/978-3-030-58610-2_6}, which detects subtle manipulative patterns by analyzing frequency aspects of images. Additionally, an ensemble model~\cite{Peng2022DFGC2T} that combines two ConvNext models trained at varying epochs and a Swin transformer has recently been proposed for enhanced deepfake detection. 
%merges three different models—two ConvNext models trained at varying epochs and a Swin-Transformer. This model leverages the strength of ConvNext's detailed feature learning and the Swin-Transformer's hierarchical structure for enhanced detection capabilities.

\subsection{Image Watermarking for Media Authentication}

The existing digital watermarking techniques are utilized to embed three types of watermarks: fragile~\cite{Martino2012FragileWT,Bhalerao2020ASI}, robust~\cite{Cox1997SecureSS,4271513,zhu2018hidden,Pereira2000RobustTM}, and semi-fragile~\cite{1180102,Lin2000DetectionOI,neekhara2022facesigns,1286417,ZHAO2024127593}. %Fragile and semi-fragile watermarks are employed to ascertain the integrity and authenticity of images. 
Fragile watermarks are particularly sensitive, designed to invalidate the authentication of an image at the slightest modification, ensuring stringent authenticity checks. In contrast, robust watermarks are crafted to endure various forms of manipulation, thus allowing content creators to assert ownership over their media, even when it undergoes alterations. Semi-fragile watermarks combine features of both fragile and robust watermarks i.e., fragile to manipulations and robust to genuine transformations. %They are used to verify if an image is authentic while also being sensitive enough to detect any manipulations made to the image.
%Semi-fragile watermarks play a critical role in the integrity and authenticity verification of images and videos, which are often compressed for transmission and storage. These watermarks are designed to survive such compression, signaling tampering only if the content is significantly altered. 
Traditional embedding techniques for semi-fragile watermarks have manipulated both the spatial, such as least significant bits~\cite{4722410}, and frequency domains (such as DCT~\cite{1286417} and DWT~\cite{Benrhouma2014TamperDA,10.1007/s11042-014-2188-7}) of digital media. 
%Spatial domain methods include direct pixel manipulation and adjustments to the least significant bits~\cite{4722410}, while frequency domain methods involve modifying frequency coefficients using techniques such as Discrete Cosine transforms (semi-fragile-DCT)~\cite{1286417} and Discrete Wavelet transforms~\cite{Benrhouma2014TamperDA,10.1007/s11042-014-2188-7}. 
However, these conventional approaches can either make watermarks perceptible, distort the media, or render them susceptible to image transformations, particularly JPEG compression. Thus, rendering them inefficient for media authentication against tampering and alteration.

%A study discussed~\cite{ZHAO2024127593} introduces a proactive approach for media authentication, utilizing deep learning to embed a \emph{semi-fragile} invisible watermark into a target image. This watermark is engineered to withstand benign image-processing operations while being susceptible to malicious tampering, serving as a verification mechanism to identify tampered regions within the image. However, this model employs a hidden secret image as the source of the watermark instead of directly embedding a bit-string within the image. Additionally, there is a lack of research investigating the impact of adversarial attacks, including both white-box and black-box attacks, on the model's performance. Furthermore, the investigation into the model's generalization to unseen manipulation techniques remains limited.

%A study in~\cite{1286417} introduces a \emph{semi-fragile} watermarking method that embeds watermark within the quantized Discrete Cosine Transform (DCT) domain of images. Although the proposed method is distinctively tolerant to JPEG compression, it is sensitive to other forms of alterations, whether they occur in the spatial or transform domain.
%This method can distinctively tolerate JPEG compression down to a predefined quality factor while being sensitive to other forms of alterations, whether they occur in the spatial or transform domains. This approach, however, is only reliable for JPEG compression; it is still ineffective for other simple transformations.
 Deep-learning-based watermarking techniques offer more efficient watermark encoding with high imperceptibility compared to traditional techniques. A robust watermarking technique called HiDDeN is proposed in~\cite{Zhu2018HiDDeNHD}, consisting of an encoder, a decoder, and a discriminator.
%A study in~\cite{Zhu2018HiDDeNHD} proposed a technique called HiDDeN, which is a \emph{robust} watermarking technique that consists of three networks: an encoder that embeds a message into a cover image, a decoder that extracts the message from the encoded image, and a discriminator network that attempts to detect whether the image is real or fake, thus enhancing the visual quality of encoded images. 
However, this technique introduces distortions in the media and is not suitable for identifying manipulated media. Similarly, a watermarking technique called StegaStamp~\cite{tancik2020stegastamp}, that encodes hyperlinks into image pixels, using a trained neural network, imperceptible to human eyes. However, this model lacks vulnerability against malicious transformations and therefore is unsuitable for media authentication.
% while retaining the perceptuality of the images.
 %It uses a trained neural network to embed hyperlinks directly into the pixel values of digital images in a way that is almost imperceptible to the human eye. These encoded images can then be printed or displayed, and the embedded data can be retrieved using a camera for media authentication.  
%aims to embed digital information invisibly into the ubiquitous imagery of the modern visual world, sidestepping the aesthetic disruption of visible QR codes. The system can encode hyperlinks into images, which remain perceptually identical to the original. These encoded images can then be printed or displayed, and the embedded data can be retrieved using a camera and the StegaStamp system. 
A study in~\cite{neekhara2022facesigns} introduces a \emph{semi-fragile} deep-learning-based invisible watermark into the image pixels (FaceSigns) utilizes a U-Net-based encoder-decoder architecture designed to be robust against benign image-processing operations yet fragile to any facial manipulation, for media authentication. However, the proposed model is not resistant to adversarial attacks based on watermark removal, rendering it unsuitable under adversarial settings.

\begin{figure*}[htbp]
\centerline{\includegraphics[width=1.0\textwidth]{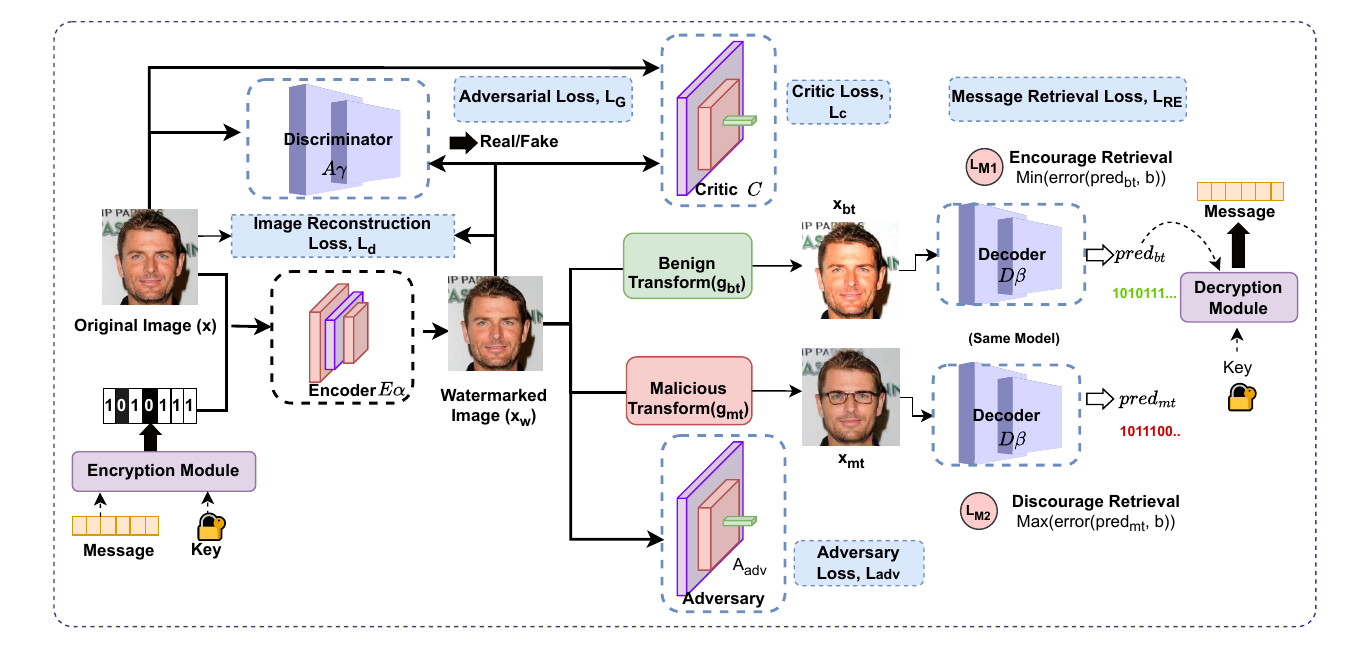}}
\caption{Overview of our proposed semi-fragile watermarking technique based on U-Net-based encoder-decoder architecture for media authentication. Training the encoder $E_{\alpha}$ and decoder $D_{\beta}$ network involves encouraging message retrieval from watermarked images that have undergone benign modifications and discouraging retrieval from watermarked images that have undergone malicious changes. The critic $C$ network is in charge of obtaining a critic score based on the quality of the image by estimating how "real" or "authentic" the images appear. The adversary network $A_{adv}$ mimics the efforts of an intruder to remove the watermark for adversarial purposes. The imperceptibility of the watermark is guaranteed by image reconstruction and adversarial loss from the discriminator $A_{\gamma}$. The loss functions proposed associated with all networks in our proposed model are also shown in the figure.
} 
\label{fig2}
\end{figure*}

%Additionally, a unique strategy~\cite{Yu2020ArtificialFF}  involves embedding watermarks directly into datasets, rather than individual images. By encoding a secret bit sequence into every image of a dataset and then training a Generative Adversarial Network (GAN) with this dataset, the watermark becomes an integral part of the images generated by the GAN. This method suggests that watermarks can be propagated through deep learning models themselves, offering a novel avenue for watermarking images generated by such models.

%Moreover, a specific application of GAN-based watermarking~\cite{9577609} has been proposed for the protection of Intellectual Property Rights (IPR) within deep generative models. This approach embeds the watermark within the model's input-output behavior and its normalization layers, aiming to prevent unauthorized replication, sharing, and redistribution. It represents a significant step towards enforcing IPR protection on GANs by ensuring that ownership can be verified through the unique watermarks embedded in the models' behaviors and outputs.

\section{Proposed Methodology}
\label{Proposed Method}

%Our goal is to create an image watermarking framework that can withstand a variety of benign image transformations as well as common generation techniques like GAN-based face swapping. 

Figure~\ref{fig2} illustrates an overview of our proposed proactive defense technique based on U-Net-based encoder-decoder architecture for invisible image watermarking. The five primary components of our proposed system are an encoder network $E_{\alpha}$, a decoder network $D_{\beta}$, an adversary network $A_{adv}$, an adversarial discriminator network $A_{\gamma}$, and a critic network $C$. 
Training the encoder $E_{\alpha}$ and decoder $D_{\beta}$  networks involves embedding watermarks and encouraging message retrieval from watermarked images that have undergone benign modifications and discouraging retrieval from watermarked images that have undergone malicious changes. The adversary network $A_{adv}$ makes an effort to mimic an intruder to remove the watermark, making it resistant to watermark removal approaches. The imperceptibility of the watermark is guaranteed by image reconstruction and adversarial loss from the discriminator $A_{\gamma}$. The critic network, denoted as $C$, is responsible for assessing the quality of images by evaluating their degree of authenticity or realism.

In detail, the encoder network $E_{\alpha}$ takes an input image $x$ and a bit string $b\in \left\{ 0,1 \right\}^{L}$
of length $L$, and outputs a watermarked image $x_{w}$ where $x_{w}=E(x,b)$. These watermarked images undergo image transformations, which include benign as well as malicious transformations. %The benign transformations are a kind of image filters which are applied to the watermarked image $x_{w}$. Similarly, malicious transforms include facial manipulation-based transformations. 
In this context, the watermarked images generated from the encoder undergo benign image transformations $(g_{bt}\sim G_{bt})$ to obtain a benign image $x_{bt}=g_{bt}(x_{w})$. Similarly, watermarked images of the encoder undergo malicious facial manipulation-based transformations $(g_{mt}\sim G_{mt})$ to obtain a malicious image $x_{mt}=g_{mt}(x_{w})$. These transformed watermarked images are fed to the decoder network to retrieve the embedded watermarked message $b_{bt}=D(x_{bt})$
and $b_{mt}=D(x_{mt})$ (note that $b^{'}$ is the notation used to denote the bit string retrieved for any image in general in the section~\ref{Evaluation criteria}), respectively.

During training, we employ the $L_{1}$ distortion between the retrieved and ground truth bit strings to optimize secret watermark retrieval. Further, the decoder is encouraged to minimize message distortion $L_{1}(b,b_{bt})$ in order to make them resilient to benign transformations, and to maximize error $L_{1}(b,b_{mt})$ to make them vulnerable to malicious manipulations. Therefore the secret retrieval error for an image $L_{RE}(x)$ is given as follows:

\begin{equation}\label{my_first_eqn}
L_{RE}(x)=L_{1}(b,b_{bt})-L_{1}(b,b_{mt})
\end{equation}

Further, we calculate the image reconstruction loss between the original $x$ and watermarked image $x_{w}$ by optimizing three specific image distortion metrics: ($L_{1}, L_{2}, L_{pips}$). Each of these metrics measures different aspects of image distortion, helping to ensure that the watermarked image retains visual fidelity to the original while embedding the necessary data. For example, the $L_{1}$ metric calculates the absolute differences between the corresponding pixel values of the original and watermarked images. Similarly, $L_{2}$, also known as the mean squared error, calculates the square of the Euclidean distance between the original and watermarked images. Finally, the $L_{pips}$ metric evaluates the perceptual similarity between two images based on their high-level features extracted from pre-trained deep networks. The pips loss is particularly effective in assessing how perceptually similar two images are, beyond just their direct pixel-wise differences. 

These metrics collectively contribute to $L_{d}(x,x_{w})$, which is used to compute the image reconstruction loss. This optimization ensures that the watermarked image closely resembles the original image in terms of aesthetics. In addition, we incorporate an adversarial loss $L_{G}(x_{w})=\log(1-A(x_{w}))$, derived from a discriminator that is concurrently trained to distinguish between the watermarked and original images. Consequently, the total image reconstruction loss is computed by combining these individual loss components.

\begin{equation}\label{my_second_eqn}
L_{d}(x,x_{w})= L_{1}(x,x_{w})+L_{2}(x,x_{w})+c_{pips}L_{pips}(x,x_{w})
\end{equation}

\begin{equation}\label{my_third_eqn}
L_{image}(x,x_{w})=L_{d}(x,x_{w})+c_{g}L_{G}(x_{w})
\end{equation}

Finally, mini-batch gradient descent is used to train the encoder and decoder network's parameters $\alpha,\beta$ to maximize the following loss over the distribution of input messages and images:

\begin{equation}\label{my_fourth_eqn}
\mathbb{E}_{x,b,g_{bt},g_{mt}}[L_{image}(x,x_{w})+c_{RE}L_{RE}(x)]
\end{equation}

Likewise, original images $x$ and watermarked images $x_{w}$ are trained using the discriminator parameters $\gamma$ as follows:

\begin{equation}\label{my_fifth_eqn}
\mathbb{E}_{x,b}[\log(1-A(x))+\log(A(x_{w}))]
\end{equation}

In the above equations, $c_{pips},c_{g},c_{RE}$ are scalar coefficients for the respective loss terms which are obtained through empirical evidence.

In addition to the encoder $E_{\alpha}$, decoder $D_{\beta}$, and discriminator networks $A_{\gamma}$, we also introduce the critic $C$ and adversary networks $A_{adv}$  in the overall model pipeline.

\subsection{Critic}
 The critic network, denoted as $C$, is responsible for assessing the quality of images by evaluating their degree of authenticity or realism. Motivating the encoder to watermark the images in a way that makes the distortion less obvious and deceives the observer, thus improving the quality of the watermarked images. The two convolutional blocks that comprise this module are followed by a linear classification layer that generates the critic score and an adaptive spatial pooling layer. The loss associated with the critic network is given as follows:

\begin{equation}\label{my_sixth_eqn}
L_{c}=\mathbb{E}_{x,b}[C(E(x,b))]
\end{equation}

 We further optimize the critic $C$ module using the Wasserstein loss function that is employed to distinguish between real and watermarked images which generally provides a more stable gradient that helps in smoother and more reliable training.

\begin{equation}\label{my_eigth_eqn}
L_{w}=\mathbb{E}_{x}[C(x)]-\mathbb{E}_{x,b}[C(E(x,b))]
\end{equation}

\subsection{Adversary}
The adversary network makes an effort to mimic an intruder to remove the watermark. To be more precise, an adversary network takes watermarked images and extracts the watermark to produce an additional set of unaffected images. This module is similar to the encoder module except that it does not have a data tensor. This module comprises two convolutional blocks followed by a linear layer that creates the residual mask. 
Subsequently, we employ a scaled TanH activation function to limit the maximum perturbation of each pixel to ±0.01. We then combine the residual mask with the watermarked image to produce the final output. The loss associated with the Adversary network is given as follows:

\begin{equation}\label{my_seventh_eqn}
L_{adv}=\mathbb{E}_{x,b}[CrossEntropy(b,D(A_{adv}(E(x,b))))]
\end{equation}

We further optimize the adversary module $A_{adv}$, which incorporates the negative cross-entropy loss to instruct the adversary to remove the embedded watermark.

\begin{equation}\label{my_ninth_eqn}
L_{r}=-\mathbb{E}_{x,b}[CrossEntropy(b,D(A_{adv}(E(x,b))))]
\end{equation}

\noindent Finally, the overall combined loss associated with the proposed model is given by:

\begin{equation} 
\label{my_tenth_eqn}
%\begin{aligned}
L_{total}=\mathbb{E}_{x,b,g_{bt},g_{mt}}[L_{image}(x,x_{w})+c_{RE}L_{RE}(x)]+\\\mathbb{E}_{x,b}[\log(1-A(x))+\log(A(x_{w}))]+L_{w}+L_{r}
%\end{aligned}    
\end{equation} 

\subsection{Message Encoding}
\label{Message Encoding}

Watermarking data is used by the encoder network as a bit string $b$ with length $L$. This watermarking data may include a secret message that may be used to verify the authenticity of the image or details about the camera that took the image. Using hashing, symmetric, or asymmetric encryption methods\footnote{https://www.enterprisenetworkingplanet.com/security/encryption-types/}, we can encrypt a target message to deter adversaries (who have obtained white-box access to the encoder network) from encoding it. In our experiments, we
incorporate $64$-bit encrypted messages, enabling the network
to encode $2^{64}$ distinct messages. Encryption involves securing data by converting readable information, termed plaintext, into an encoded format called ciphertext. In our study, we employ symmetric encryption, where the same secret key is used for both the encryption and the decryption processes. Specifically, we utilize the Data Encryption Standard (DES), a symmetric key algorithm designed for electronic data encryption. DES functions as a block cipher, encrypting data in blocks, typically $64$-bit blocks, using a $56$-bit secret key for message encryption or decryption~\cite{schneier1996applied}. %Notably, DES represents a significant advancement in cryptographic algorithm development and has influenced the evolution of other encryption technologies.

\subsection{Network Architectures}

Following existing studies in~\cite{isola2017image,neekhara2022facesigns}, the foundation of our encoder and decoder networks is the U-Net convolutional neural network architecture which takes images of size $224\times 224$. Initially, a fully trained trainable layer converts encrypted messages, which are represented as an L-length bitstring, to the $84\times84$ tensor $b_{proj}$. The original RGB image is scaled to $224\times224$ using bilinear interpolation, and these tensors are then added as the fourth channel to form the encoder network's input. There are eight downsampling and eight upsampling layers in the U-Net encoder. As recommended by ~\cite{46191,neekhara2022facesigns}, we improve the original U-Net architecture by substituting convolutions followed by nearest-neighbor upsampling for transposed convolutions in the upsampling layers. The structure of the decoder network substantially resembles the encoder network. First, the U-Net decoder creates an intermediate output that is $224\times 224$. The bilinear downsampling is then used to reduce the size of the intermediate output to $84\times84$, creating $b_{Decoded}$. After that, a fully connected layer project $b_{Decoded}$ onto a vector of size $L$. A sigmoid layer is then used to scale the values between $0$ and $1$.

We used the patch discriminator described in~\cite{isola2017image} for the discriminator network. The discriminator's job is to identify if each $N\times N$ image patch is legitimate or fake. To obtain the output of the discriminator, we aggregate the discriminator responses across all patches. Three convolutional blocks with a stride of $2$ are used for our discriminator network, making it easier to classify patches of size $28\times28$.  \\

\noindent \textbf{Transformation Functions:} 
In our work, we used benign and malicious transformation functions to establish the robustness and fragility of the embedded watermark using our proposed model.

%The spectrum of possible benign and malicious image modifications in real-world circumstances is extensive given the fact that we can only utilize just a handful of them during training. 

%Through our studies, we demonstrate that generalization to previously unseen benign and malicious transformations can be achieved by incorporating the transformation functions below.
\noindent \textit{Benign Transforms:} During training, we apply the diverse set of differentiable benign image transformations, denoted as ($G_{bt}$), to our watermarked images, in order to imitate usual image processing operation.

 \begin{enumerate}
\item \noindent \textbf{JPEG Compression:} Recall that during training, we apply JPEG compression with quality 25\%, 50\%, and 75\%. We use the differentiable JPEG function introduced in~\cite{Shin2017JPEGresistantAI} to approximate JPEG compression. 
\item \noindent \textbf{Gaussian Blur:} We use a Gaussian kernel $k$ to convolve the original image. The expression for this transform is t(x) = k * x, where $*$ denotes the convolution operator. We employ kernel sizes between $k = 5$ and $k = 10$.
\item \noindent \textbf{Saturation Settings:} We randomly linearly interpolate between the original image and its grayscale version to allow for different color modifications from social media filters.
\item \noindent \textbf{Contrast Settings:} Using a contrast factor $\sim u[0.8,1.8]$, we linearly rescale the histogram of the image.
\item \noindent \textbf{Downsizing and Upsizing:} Using bilinear upsampling, the image is first downscaled by a factor scale $\sim u[3,8]$ and then upsampled by the same factor.
\item \noindent \textbf{Translation and rotation:} We shift image both horizontally and vertically by $n_{h}$ and $n_{v}$ pixels where $n_{h},n_{v}\sim u[-8,8]$ and rotated by $\theta$ degrees where $\theta\sim u[-8,8]$
\end{enumerate}
In general, Compression attacks, like those using JPEG algorithms, are deliberate attempts to degrade image quality or test system vulnerabilities, often introducing artifacts to obscure details. In contrast, social media platforms apply compression to optimize storage and improve loading times, aiming for efficient data use while maintaining acceptable visual quality. While both processes involve lossy compression, the former is typically used for exploiting or testing, whereas the latter is a routine practice for user experience.

During training, we selected one transformation function from the aforementioned list, together with an Identity transform, for every mini-batch iteration, and we applied it to every image in the batch. \\

\noindent \textit{Malicious Transforms:} The embedded watermarks obtained using our proposed model should be vulnerable to all malicious attacks or generative techniques. To facilitate this, we assume that all Deepfake approaches operate by modifying facial features to mimic the appearance of the target identity. Consequently, we represent malicious manipulation as a transformation function ($g_{mt}$) that specifically involves changing the watermark within certain facial regions. To this front, the facial landmarks are detected using MTCNN~\cite{zhang2016joint}, Then these points are used as vertices to create polygons. For example, the landmarks identifying the outline of the lips are connected to form a lip polygon. We used the Dlib library~\cite{10.5555/1577069.1755843} to draw these polygons on the image by connecting the landmark points. This library provides functions to draw shapes based on specified points. Then, for every image, we create a mask $M_{h\times w\times c}$ made up of all ones. Then, we locate the polygons that represent the lips, nose, and eyes on the face, and we set the pixel values inside these polygons to a preset watermark retention percentage, $w_{r}\in [0,1]$. In other words, M[i, j, :] = $w_{r}$ for all pixels (i, j) inside the face feature polygons. Ultimately, the maliciously altered image, $g_{mt}(x_{w})$, is determined based on the below equation:

\begin{equation}\label{my_eleventh_eqn}
g_{mt}(x_{w})= M.x_{w}+(1-M).x
\end{equation}

\noindent Thus, based on the aforementioned configuration, we have implemented \textbf{three} different versions of this model as shown in Table~\ref{TableConfig} where U-Net denotes the encoder-decoder model with the adversarial network, $C$ is the critic model, and $A_{adv}$ is the adversarial network.

\begin{comment}
  \begin{enumerate}
\item U-Net+Critic+Adversary (Baseline) without any benign and malicious transformations used during the training stage.
\item U-Net+Critic+Adversary when only benign transformations ($g_{bt}$) are used during training.
\item U-Net+Critic+Adversary when both benign transformations ($g_{bt}$) and malicious transformations ($g_{mt}$) are used during training.
\end{enumerate}   
\end{comment}

\begin{figure}[htbp]
\centerline{\includegraphics[width=0.35\textwidth]{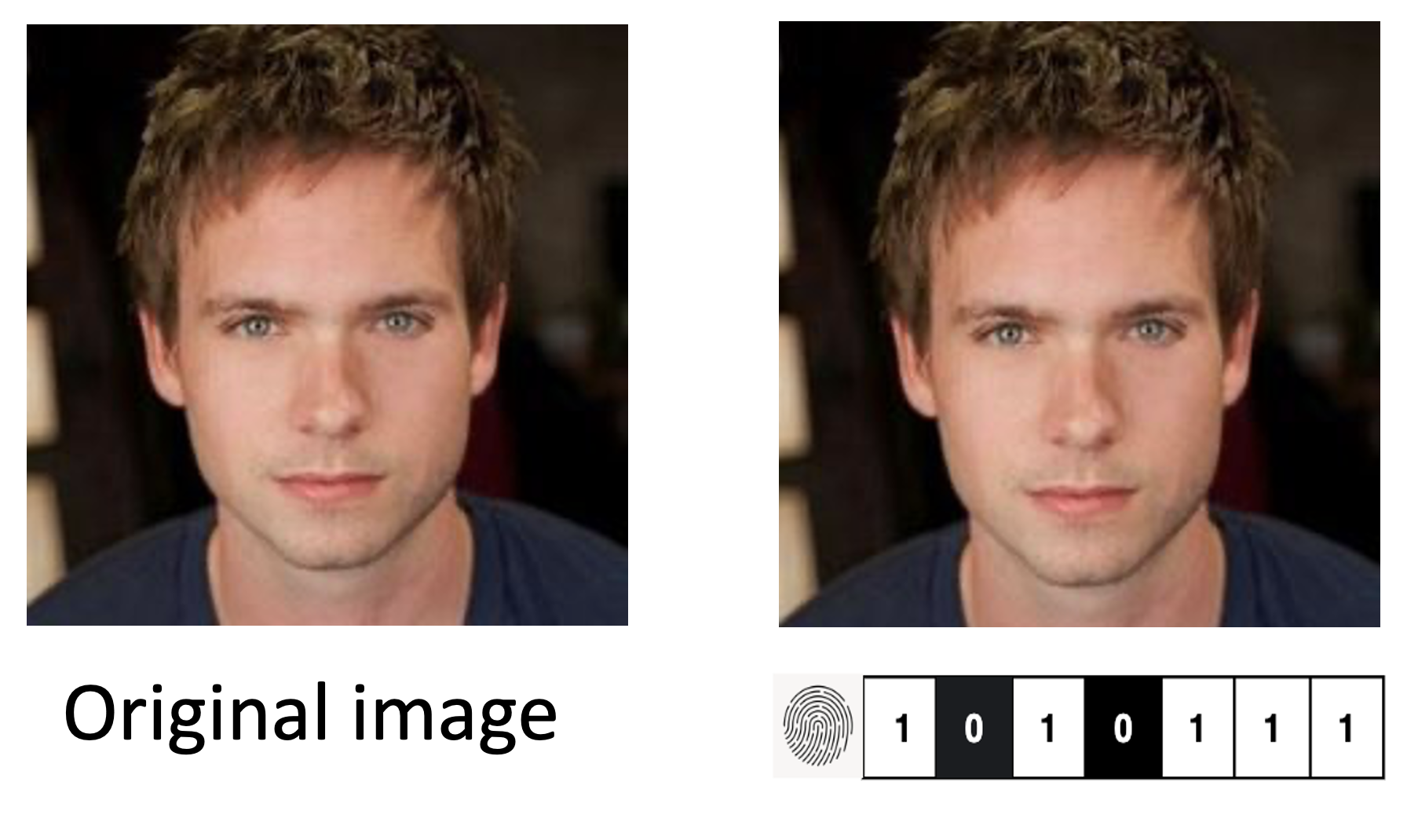}}
\caption{Pictorial representation of watermarked output $x_{w}$ when original image $x$ is given to our proposed model for watermarking.}
\label{fig3}
\end{figure}

\begin{comment}

\begin{figure}[htbp]
\centerline{\includegraphics[width=0.5\textwidth]{Fig 3.png}}
\caption{Sample (a) real  and (b) fake images from the FF++ dataset.}
\label{fig4}
\end{figure}

\begin{figure}[htbp]
\centerline{\includegraphics[width=0.5\textwidth]{Fig 4.png}}
\caption{Sample Real Images from Celeb A dataset.}
\label{fig5}
\end{figure}

\begin{figure}[htbp]
\centerline{\includegraphics[width=0.5\textwidth]{Fig 5.png}}
\caption{Sample Real Images from IMDB-WIKI dataset.}
\label{fig6}
\end{figure}
\end{comment}

\begin{figure*}[htbp]
\centerline{\includegraphics[width=1.0\textwidth]{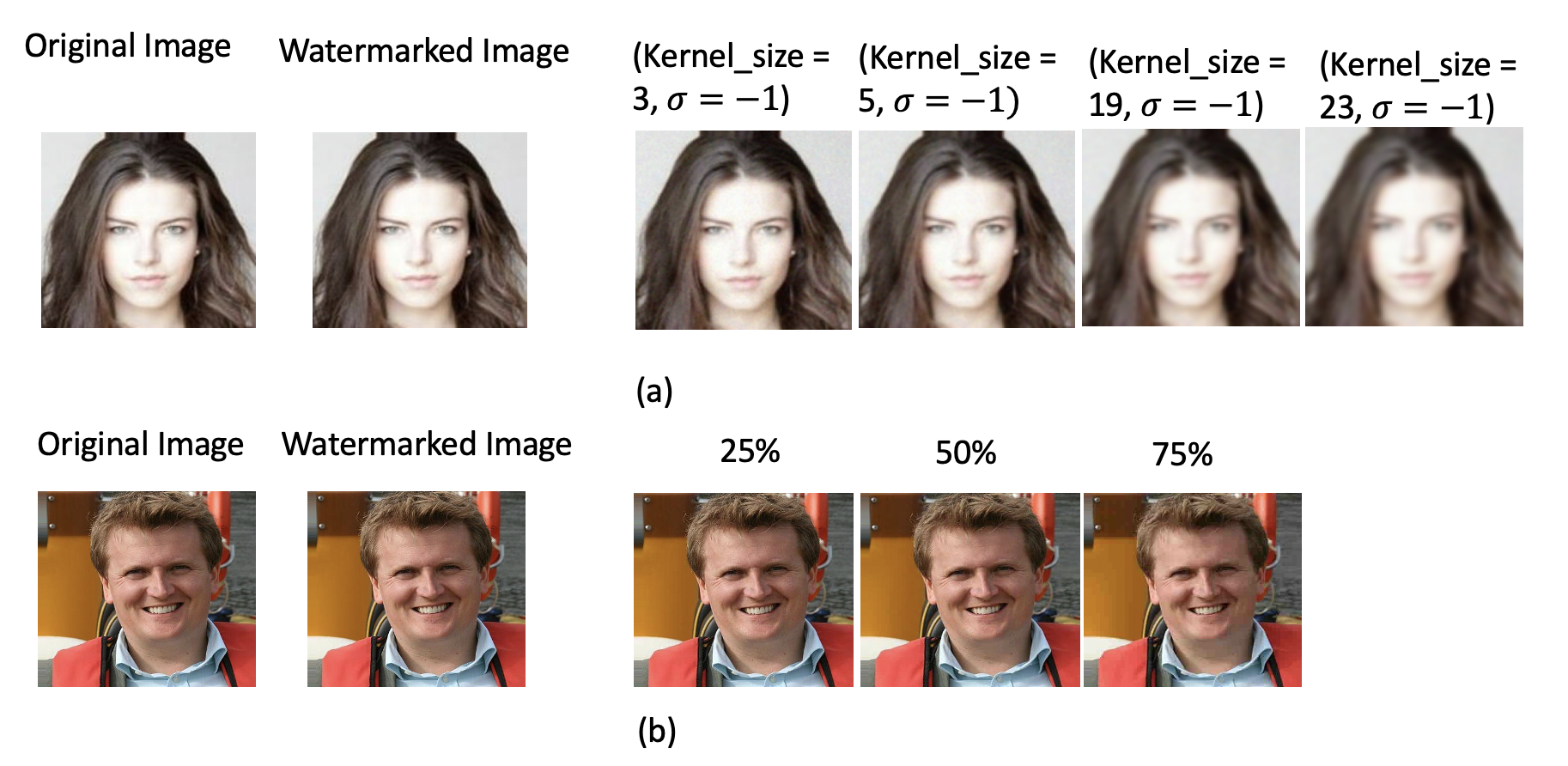}}
\caption{(a) Illustration of Gaussian blur on invisible watermarked images using different kernel sizes and the $\sigma$ values. 
 (b) Application of JPEG compression to invisible watermarked images at different compression rates ranging from 25 → 75 [best viewed in Zoom].
} 
\label{Gauss_JPEG}
\end{figure*}

\begin{figure*}[htbp]
\centerline{\includegraphics[width=0.9\textwidth]{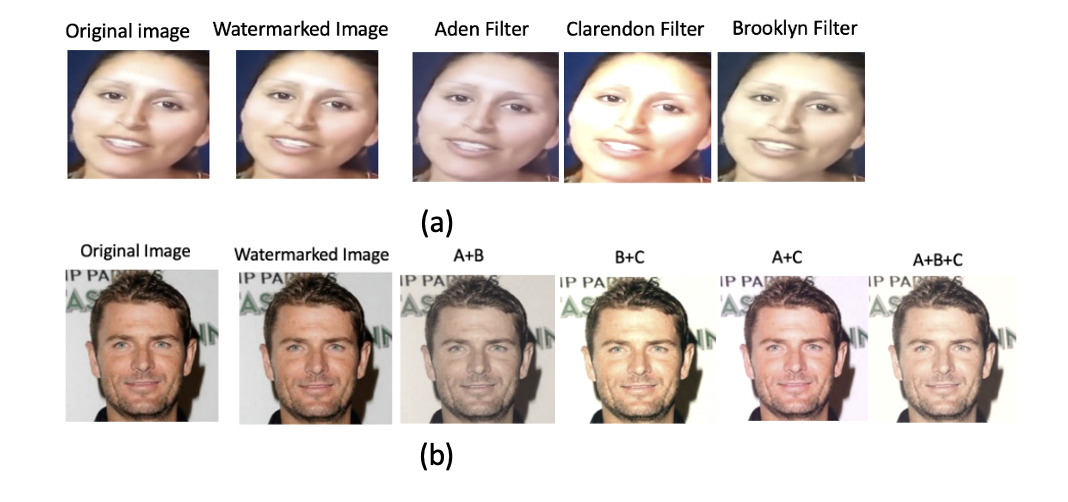}}
\caption{(a) Application of unseen benign transforms on invisible watermarked images. Instagram filters like Brooklyn, Clarendon, and Aden are examples of benign transformations shown in this diagram. (b) Combined application of unseen benign transforms on invisible watermarked images. In this work, Instagram filters such as Brooklyn, Clarendon, and Aden are used. The symbol (A+B) in the figure denotes the combined application of the Brooklyn and Aden filters to the watermarked image. Similarly, (B+C) in the figure denotes the combined application of Brooklyn and Clarendon filters to the watermarked image. Finally, (A+B+C) in the figure denotes the combined application of Aden, Brooklyn, and Clarendon filters to the watermarked image.
} 
\label{benign_filters}
\end{figure*}

%Thus, based on aforementioned configuration, we have implemented three different versions of this model as shown in table~\ref{TableConfig}.

\begin{table}
\caption{Implementation of different configurations of our proposed model.}
\label{TableConfig}
\begin{center}
\scalebox{0.84}{
\begin{tabular}{l|c}
\hline
\multicolumn{1}{c|}{Model} & Transformations used during training                                                    \\ \hline
Our-U-Net+C+$A_{adv}$ (Baseline)     & No Transformations                                                           \\
Our-U-Net+C+$A_{adv}$ ($g_{bt}$)   & Only Benign Transformations                                           \\
Our-U-Net+C+$A_{adv}$ ($g_{bt}$,$g_{mt}$)     & Both Benign and Malicious Transformations \\ \hline
\end{tabular}}
\end{center}
\end{table}

\section{Experimental Validations}
\label{Experimental validations}

In our experiments, we used three datasets, namely, FaceForensics++, CelebA, and IMDB-WIKI datasets for both intra-dataset and cross-dataset evaluations.
\subsection{Datasets}
\begin{itemize}
  \item \textbf{FaceForensics++:}  FaceForensics++ (FF++)~\cite{9010912} is an automated benchmark for facial manipulation detection. It consists of several manipulated videos created using two different generation techniques: Identity Swapping (FaceSwap, FaceSwap-Kowalski, FaceShifter, Deepfakes) and Expression Swapping (Face2Face and NeuralTextures). We used the FF++ dataset's $c23$ version for both training and testing ($80\%$ videos for training, $20\%$ videos for testing, with 60 frames per video). We used the real images from this dataset for embedding watermarks using our model. %Fig.~\ref{fig4} shows the sample real and fake images from the FF++ dataset.

  \item \textbf{CelebA:} The Large-scale CelebFaces Attributes Dataset (CelebA)~\cite{liu2015faceattributes} is the publicly available face dataset with more than $200K$ celebrity images. % from $10K$ subjects with $40$ soft 
  In addition, this dataset covers large pose variations and background clutter with $10k$ identities, $202,599$ face images, $5$ landmark locations, and $40$ binary attribute annotations per image. This dataset is used to train our model ($70\%$ used for training and $30\%$ used for testing) to generate watermarked facial images. %Figure~\ref{fig5} shows the sample real images from the CelebA dataset.

  \item \textbf{IMDB-WIKI:} IMDB-WIKI~\cite{7406390} is a highly curated dataset of popular celebrities that is created from both the IMDB website and Wikipedia. The dataset has rich annotations like DOB, year of photo taken, gender, name of the celebrity, and celebrity ID along with other essential information in the metadata file. Altogether, the dataset has $460,723$ facial images representing $20,284$ celebrities sourced from IMDb, and $62,328$ images from Wikipedia, resulting in a combined total of $523,051$ images. %Figure~\ref{fig6} shows the sample real images from the IMDB-WIKI dataset.
\end{itemize}
\subsection{Training Procedure} 
The training process involves $70,000$ mini-batch iterations of images with a batch size of $64$, employing an Adam optimizer with a fixed learning rate of $0.0001$. All the implementation was performed using Python, and all the models are trained using images of size $224\times224$, obtained after cropping and resizing the images from the data sets. In our experiments, we consider a message length of L=$64$ (i.e. $2^{6}$) bits. Here, the bit length is explicitly specified as L=$64$, which is equivalent to $2^{6}$ bits. This means that each message handled in these experiments consists of $64$ binary digits (bits), allowing a substantial amount of information per message given that the total number of possible distinct messages that can be created with $64$ bits is $2^{64}$. This message length is quite significant because it offers a large combination of bits, enabling strong encryption standards or the capacity to handle complex data structures or identifiers in computational and cryptographic applications.  The parameters are computed once for the
model during the offline training stage, and not for each input image. Figure~\ref{fig3} shows the pictorial representation of watermarked output $x_{w}$ when the original image $x$ is input to our proposed model for watermarking.

We primarily evaluate image watermark embedding techniques based on the following criteria:

\subsection{Evaluation Criteria}
\label{Evaluation criteria}

\begin{itemize}
  \item \textbf{Imperceptibility and Capacity:}
   For imperceptibility, we compare the original and watermarked images using two metrics: Peak Signal-to-Noise Ratio~(PSNR) and Structural Similarity Index~(SSIM). Both PSNR and SSIM are widely used to assess the quality and perceptibility of watermarked images. A higher PSNR signifies less distortion between the original and watermarked images, while a higher SSIM value indicates closer resemblance. Thus, higher values for both metrics are preferable, suggesting a more imperceptible watermark.

  Capacity refers to the quantity of information that can be 
  successfully embedded within an image. This metric is crucial in contexts like digital watermarking, where you need to hide data within visual content without affecting the perceptibility or integrity of the original image. The amount of bits of the encrypted message embedded in each pixel of the image is measured in terms of bits per pixel, or BPP which corresponds to the number of bits of the encrypted message embedded per pixel of the image. This is simply calculated as the ratio of the message length (L) to the total number of pixels in the image ($H\times W\times C$) defining the capacity. 
  
  %The trade-off between imperceptibility and capacity should be noted. For instance, increasing the capacity, or the amount of data that can be embedded within an image often means altering more pixels or making more significant changes to the image data. %This allows more information to be stored by potentially reducing imperceptibility, making the modifications more noticeable to an observer. 
  The challenge is to embed enough data without impacting the imperceptibility of the image.\\

\item \textbf{Robustness and Fragility:}
We quantify the Bit Recovery Accuracy (BRA) of the embedded watermarked images under the unknown benign and malicious image transformations to evaluate the robustness and fragility of the watermarking technique. A high Bit Recovery Accuracy (BRA) against unseen benign modifications would indicate that the watermarked or embedded data should remain detectable and recoverable even after the image has undergone common image processing operations. On the other hand, a low BRA is preferred for fragility against malicious transformations such as Deepfakes. 

In this context, to calculate the Bit Recovery Accuracy (BRA), we directly compare the original input bit string, denoted $b$, with the decoded output, 
$b^{'}$, from the decoder. Recall that in our experiments, we use the same $56$-bit secret key for both encryption and decryption of $64$-bit message bit string using Data encryption standard~(DES)~\cite{schneier1996applied} (please refer to section~\ref{Message Encoding}). This $56$-bit secret key is used to decrypt the extracted $64$-bit string ($b^{'}$) from the watermarked image.

Let $n$ represent the total number of bits in $b$ (and $b^{'}$, assuming they have the same length), and let $m$ denote the number of bits that match between $b$ and $b^{'}$ using a distance metrics, hamming distance in our study. The BRA is then calculated using the following equation:

\begin{equation}\label{my_BRA}
BRA = \frac{m}{n}\times 100\%
\end{equation}

\end{itemize}

 %As the capacity increases, the complexity of the data embedded can also increase, potentially complicating the process of accurately recovering the embedded information. High-capacity embedding might use more complex or dense encoding schemes, which can be more susceptible to errors during extraction, especially if the image undergoes compression or other forms of degradation after embedding.

% Similarly, models that focus on robustness might use more redundant or complex encoding methods to protect the embedded data, which can reduce the total capacity available for embedding additional data. The strength of the encryption or the robustness of the watermark might take up more space, leaving less room for actual data.
\section{Imperceptibility and Capacity}
\label{Imperceptibility and Capacity}
%\noindent \textbf{Imperceptibility and Capacity:} 
We evaluate the imperceptibility and capacity of our proposed watermarking framework against four existing deep-learning-based image watermarking techniques: HiDDeN~\cite{zhu2018hidden}, StegaStamp~\cite{tancik2020stegastamp}, Semi-fragile-DCT~\cite{1286417}, and FaceSigns~\cite{neekhara2022facesigns} trained on benign and malicious transformations. %All these models are trained and tested on 
%While,  HiDDeN and StegaStamp both successfully embed a bitstring message into a square RGB image and ensure it withstands various image transformations, their approach is different from the semi-fragile DCT watermarking method. 
%Semi-fragile DCT watermarking is specifically designed to be resilient to JPEG compression, allowing it to maintain integrity when an image undergoes this common form of data compression. However, it remains vulnerable to other image transformations such as Gaussian blur, scaling, and rotation. 

\begin{table}
\caption{Capability and imperceptibility measures of various
invisible image watermarking schemes. The input image’s width and height
are denoted by $H$ and $W$.}
\label{Table1}
\begin{center}
\scalebox{1.0}{
\begin{tabular}{l|ccc|cc}
\hline
\multicolumn{1}{c|}{\multirow{2}{*}{Method}} & \multicolumn{3}{c|}{Capacity} & \multicolumn{2}{l}{Imperceptibility} \\ \cline{2-6} 
\multicolumn{1}{c|}{}                        & H,W     & L      & BPP        & PSNR              & SSIM             \\ \hline
SemiFragile-DCT                              & 128     & 256    & 5.2e-3     & 20.29             & 0.846            \\
Hidden                                       & 128     & 30     & 6.1e-4     & 24.96             & 0.928            \\
StegaStamp                                   & 400     & 100    & 2.0e-4     & 28.64             & 0.922            \\                     
FaceSigns (Semi-Fragile)                     & 256     & 128    & 6.5e-4     & \textbf{36.08}
   & \textbf{0.975}            \\ \hline
Our-U-Net (Baseline)                             & 224     & 64     & 3.71e-4    & 34.24             & 0.959           \\
Our-U-Net+C                                & 224     & 64     & 3.71e-4    & 34.82             & 0.965            \\
Our-U-Net+C+$A_{adv}$                      & 224     & 64     & 3.71e-4    & \textbf{35.57}             & \textbf{0.970}           \\ \hline
\end{tabular}}
\end{center}
\end{table}

In Table~\ref{Table1}, we present the image imperceptibility and capacity metrics of different watermarking techniques. Our observations reveal that our model achieves superior imperceptibility in encoding messages compared to those encoded by StegaStamp, HiDDeN, and Semi-fragile-DCT, as denoted by higher PSNR and SSIM. The imperceptibility of our proposed model is at par with the FaceSigns model (trained on both benign and malicious transformations) which is due to the similar backbone architecture i.e., U-Net.
We also observed that while it's commonly believed that there is a trade-off between capacity and imperceptibility, this isn't always the case. For instance, as shown in Table~\ref{Table1}, the HiDDeN watermarking scheme with just a $30$-bit length did not yield better imperceptibility in terms of PSNR and SSIM. This suggests that the quality of imperceptibility depends significantly on how effectively the bit string is embedded, rather than solely on the capacity of the data embedded.

The improved performance of our model is largely due to the difference in the network architecture. In addition, we utilized an intermediate message reconstruction loss which encourages the network to preserve important features and details throughout its depth, which might otherwise be lost during downsampling in the contracting path.
Furthermore, our model (Our-U-Net+C+$A_{adv}$) employs nearest-neighbor upsampling instead of transposed convolutions. This choice helps minimize upsampling artifacts, further improving the imperceptibility of the image with an embedded watermark, as also noted in~\cite{neekhara2022facesigns}. We also did experiments with a $128$-bit length string using our proposed model and obtained similar imperceptibility as compared to a $64$-bit length string. Therefore, we used a $64$-bit length string for all the remaining experiments as a more computational-friendly alternative.
\\

\section{Robustness and Fidelity}
\label{Robustness and Fidelity}
In order to examine the resilience and susceptibility of different Deep Neural Network (DNN)-based watermarking methods, we expose the watermarked images to unseen benign and malicious transformations and evaluate the retrieved BRA. The results highlighted in bold are the top performances.

\subsection{Benign Transform}
  %Next, we try to extract the embedded message from the manipulated images. Thus, we analyze practical and inaccessible changes, usually performed before posting images to the public domain.

For benign transformations, we applied different levels of Gaussian blur, JPEG compression, and different Instagram filters, namely, Aden, Brooklyn, and Clarendon. We utilized the open-source Pilgram library~\cite{https://github.com/akiomik/pilgram} to implement various Instagram filters, including Aden, Brooklyn, and Clarendon, to test their impact on our proposed watermarking scheme. This Pilgram~\cite{https://github.com/akiomik/pilgram} library provides a range of image processing filters, including Instagram-style filters. Figure~\ref{Gauss_JPEG} (a) shows the illustration of Gaussian blur on invisible watermarked images using different kernel sizes and $\sigma$ values. Figure~\ref{Gauss_JPEG} (b) shows the illustration of JPEG compression on invisible watermarked images at different compression rates varying from $25\% \to 75$\%, respectively. For the Figure~\ref{Gauss_JPEG}, the watermarked samples are generated using our U-Net+C+$A_{adv}$ model trained only on benign transformations.

\begin{table}[!htbp]
\caption {The effect of Gaussian Blur on the invisible watermarked images, in terms of Bit recovery accuracy (BRA) with different kernel sizes and $\sigma$ values on different versions of our proposed model as shown in Table~\ref{TableConfig}.} %(baseline), proposed model when only benign transformations $g_{bt}$ are used during training and proposed model when both benign $g_{bt}$ and malicious transformations $g_{mt}$ are used during training.}
\label{Table2}
\begin{center}
\scalebox{0.70}{
\begin{tabular}{l|lcccc}
\hline
\multicolumn{1}{c|}{\multirow{2}{*}{Method}} & \multicolumn{5}{c}{Gaussian Blur (BRA$\%$)}                                                                                                                                                                                                                                                              \\ \cline{2-6} 
\multicolumn{1}{c|}{}                        & \multicolumn{1}{c|}{None}           & \multicolumn{1}{l|}{(Kernel\_size=3,\textbackslash{}sigma=-1)} & \multicolumn{1}{l|}{(Kernel\_size=5,\textbackslash{}sigma=-1)} & \multicolumn{1}{l|}{(Kernel\_size=19,\textbackslash{}sigma=-1)} & \multicolumn{1}{l}{(Kernel\_size=23,\textbackslash{}sigma=-1)} \\ \hline
SemiFragile DCT                              & \multicolumn{1}{l|}{99.43}          & \multicolumn{1}{c|}{76.24}                                     & \multicolumn{1}{c|}{72.19}                                     & \multicolumn{1}{c|}{67.63}                                      & 63.49                                                          \\
Hidden                                       & \multicolumn{1}{l|}{97.65}          & \multicolumn{1}{c|}{84.97}                                     & \multicolumn{1}{c|}{82.29}                                     & \multicolumn{1}{c|}{79.45}                                      & 73.62                                                          \\
StegaStamp                                   & \multicolumn{1}{l|}{\textbf{99.62}} & \multicolumn{1}{c|}{99.14}                                     & \multicolumn{1}{c|}{98.65}                                     & \multicolumn{1}{c|}{94.95}                                      & 92.48                                                          \\
FaceSigns (Semi-Fragile)                     & \multicolumn{1}{l|}{99.49}          & \multicolumn{1}{c|}{98.24}                                     & \multicolumn{1}{c|}{96.58}                                     & \multicolumn{1}{c|}{94.12}                                      & 91.65                                                          \\ \hline
U-Net+C+$A_{adv}$ (Baseline)                 & \multicolumn{1}{l|}{99.29}          & \multicolumn{1}{c|}{95.56}                                     & \multicolumn{1}{c|}{92.61}                                     & \multicolumn{1}{c|}{89.37}                                      & 85.95                                                          \\
U-Net+C+$A_{adv}$ ($g_{bt}$)                 & \multicolumn{1}{l|}{99.45}          & \multicolumn{1}{c|}{\textbf{99.18}}                            & \multicolumn{1}{c|}{\textbf{98.76}}                            & \multicolumn{1}{c|}{\textbf{96.59}}                             & \textbf{93.87}                                                 \\
U-Net+C+$A_{adv}$ ($g_{bt}$+$g_{mt}$)        & \multicolumn{1}{l|}{99.31}          & \multicolumn{1}{c|}{98.84}                                     & \multicolumn{1}{c|}{97.28}                                     & \multicolumn{1}{c|}{95.58}                                      & 92.26                                                          \\ \hline
\end{tabular}}
\end{center}
\end{table}

Table~\ref{Table2} tabulates the BRA in $\%$ for different watermarking techniques after applying Gaussian blur as a benign transform at different kernel sizes and $\sigma$ values. 
Table~\ref{Table3} tabulates BRA in $\%$ for different watermarking techniques after applying JPEG compression (benign transform) at compression rates of 25$\%$, 50$\%$ and 75$\%$, respectively. Table~\ref{Table4} tabulates the effect of unseen benign transformations on invisible watermarked images in terms of Bit Recovery Accuracy (BRA) using different Instagram filters. All models are trained and tested on the CelebA dataset.

The overall performance in terms of BRA is $97.1\%$ for the U-Net+C+$A_{adv}$ on Gaussian blur which is better than $96.3\%$ BRA of the second-best model, StegaStamp, when only benign transformations $g_{bt}$ are used during the training stage. Similar observations were seen on JPEG compression and unseen benign transformations. %The reason is that when a model is exclusively trained using benign transformations, such as cropping, compression, or subtle filtering, it becomes highly adept at recognizing and handling the specific patterns and distortions associated with these types of transformations. Consequently, the model demonstrates excellent performance when evaluated on similar benign transformations.

%across different kernel sizes and $\sigma$ values. The reason is that when a model is exclusively trained using benign transformations, such as cropping, compression, or subtle filtering, it becomes highly adept at recognizing and handling the specific patterns and distortions associated with these types of transformations. This specialized training equips the model to be resilient to the similar kinds of transformations, that it was exposed to, during its training phase. As a result, it performs exceptionally well on similar benign transformations.

 %The overall performance in terms of BRA is better for the U-Net+C+$A_{adv}$ when only benign transformations $g_{bt}$ are used during training. The consistent performance across various compression rates can be attributed to the model's training regimen, which focuses solely on benign transformations like mild cropping, slight compression, or subtle filtering. When a model is trained exclusively with these types of transformations, it becomes highly proficient at identifying and dealing with the specific patterns and distortions associated with them. This specialized training prepares the model to effectively handle the same kinds of noise or alterations it encountered during training. Consequently, the model demonstrates excellent performance when faced with similar benign transformations.
\begin{table}
\caption{The effect of JPEG Compression on the invisible watermarked images in terms of Bit Recovery Accuracy (BRA) at different compression rates on different versions of our proposed model as described in Table~\ref{TableConfig}.}% (baseline), proposed model when only benign transformations $g_{bt}$ are used during training and proposed model when both benign $g_{bt}$ and malicious transformations $g_{mt}$ are used during training.}
\label{Table3}
\begin{center}
\scalebox{1.0}{
\begin{tabular}{l|lccc}
\hline
\multicolumn{1}{c|}{\multirow{2}{*}{Method}} & \multicolumn{4}{c}{JPEG Compression (BRA$\%$)}                                                                                               \\ \cline{2-5} 
\multicolumn{1}{c|}{}                        & \multicolumn{1}{c|}{None}           & \multicolumn{1}{l|}{25$\%$}         & \multicolumn{1}{l|}{50$\%$}         & \multicolumn{1}{l}{75$\%$} \\ \hline
SemiFragile DCT                              & \multicolumn{1}{l|}{99.43}          & \multicolumn{1}{c|}{60.13}          & \multicolumn{1}{c|}{58.28}          & 53.86                      \\
Hidden                                       & \multicolumn{1}{l|}{97.65}          & \multicolumn{1}{c|}{71.45}          & \multicolumn{1}{c|}{70.64}          & 67.19                      \\
StegaStamp                                   & \multicolumn{1}{l|}{\textbf{99.62}} & \multicolumn{1}{c|}{98.38}          & \multicolumn{1}{c|}{\textbf{97.28}} & \textbf{95.24}             \\
FaceSigns (Semi-Fragile)                     & \multicolumn{1}{l|}{99.49}          & \multicolumn{1}{c|}{98.54}          & \multicolumn{1}{c|}{95.38}          & 93.75                      \\ \hline
U-Net+C+$A_{adv}$ (Baseline)                 & \multicolumn{1}{l|}{99.29}          & \multicolumn{1}{c|}{92.49}          & \multicolumn{1}{c|}{77.78}          & 71.59                      \\
U-Net+C+$A_{adv}$ ($g_{bt}$)                 & \multicolumn{1}{l|}{99.45}          & \multicolumn{1}{c|}{\textbf{98.68}} & \multicolumn{1}{c|}{95.87}          & 94.52                      \\
U-Net+C+$A_{adv}$ ($g_{bt}$+$g_{mt}$)        & \multicolumn{1}{l|}{99.31}          & \multicolumn{1}{c|}{96.68}          & \multicolumn{1}{c|}{92.25}          & 89.45                      \\ \hline
\end{tabular}
}
\end{center}
\end{table}

\begin{table}[!htbp]
\caption {The effect of unseen benign transformations on the invisible watermarked images in terms of Bit recovery accuracy (BRA) when different Instagram filters are applied on the watermarked images obtained using different versions of our proposed model as described in Table~\ref{TableConfig}.}
\label{Table4}
\begin{center}
\scalebox{0.69}{
% Please add the following required packages to your document preamble:
% \usepackage{multirow}
\begin{tabular}{l|lccccccc}
\hline
\multicolumn{1}{c|}{\multirow{2}{*}{Method}} & \multicolumn{8}{c}{Instagram Filters (BRA$\%$)}                                                                                                                                                                                                                         \\ \cline{2-9} 
\multicolumn{1}{c|}{}                        & \multicolumn{1}{c|}{None}  & \multicolumn{1}{l|}{Aden}  & \multicolumn{1}{l|}{Brooklyn} & \multicolumn{1}{l|}{Clarendon} & \multicolumn{1}{c|}{Aden+Brooklyn} & \multicolumn{1}{c|}{Brooklyn+Clarendon} & \multicolumn{1}{c|}{Aden+Clarendon} & Aden+Brooklyn+Clarendon \\ \hline
SemiFragile DCT                              & \multicolumn{1}{l|}{99.43} & \multicolumn{1}{c|}{93.47} & \multicolumn{1}{c|}{95.79}    & \multicolumn{1}{c|}{95.12}     & \multicolumn{1}{c|}{92.39}         & \multicolumn{1}{c|}{93.42}              & \multicolumn{1}{c|}{94.74}          & 91.56                   \\
Hidden                                       & \multicolumn{1}{l|}{97.65} & \multicolumn{1}{c|}{94.61} & \multicolumn{1}{c|}{93.82}    & \multicolumn{1}{c|}{94.13}     & \multicolumn{1}{c|}{91.37}         & \multicolumn{1}{c|}{91.29}              & \multicolumn{1}{c|}{92.23}          & 89.79                   \\
StegaStamp                                   & \multicolumn{1}{l|}{\textbf{99.62}} & \multicolumn{1}{c|}{\textbf{99.48}} & \multicolumn{1}{c|}{\textbf{99.26}}    & \multicolumn{1}{c|}{99.09}     & \multicolumn{1}{c|}{97.18}         & \multicolumn{1}{c|}{96.28}              & \multicolumn{1}{c|}{95.79}          & 94.13                   \\
FaceSigns (Semi-Fragile)                     & \multicolumn{1}{c|}{99.49} & \multicolumn{1}{c|}{99.45} & \multicolumn{1}{c|}{99.22}    & \multicolumn{1}{c|}{99.15}     & \multicolumn{1}{c|}{98.53}         & \multicolumn{1}{c|}{97.86}              & \multicolumn{1}{c|}{97.51}          & 96.36                   \\ \hline
U-Net+C+$A_{adv}$ (Baseline)                 & \multicolumn{1}{l|}{99.29} & \multicolumn{1}{c|}{99.27} & \multicolumn{1}{c|}{98.87}    & \multicolumn{1}{c|}{98.64}     & \multicolumn{1}{c|}{97.24}         & \multicolumn{1}{c|}{96.02}              & \multicolumn{1}{c|}{95.09}          & 94.61                   \\
U-Net+C+$A_{adv}$ ($g_{bt}$)                 & \multicolumn{1}{l|}{99.45} & \multicolumn{1}{c|}{99.39} & \multicolumn{1}{c|}{99.25}    & \multicolumn{1}{c|}{\textbf{99.19}}     & \multicolumn{1}{c|}{\textbf{98.87}}         & \multicolumn{1}{c|}{\textbf{98.54}}              & \multicolumn{1}{c|}{\textbf{98.69}}          & \textbf{97.18}                   \\
U-Net+C+$A_{adv}$ ($g_{bt}$+$g_{mt}$)        & \multicolumn{1}{l|}{99.31} & \multicolumn{1}{c|}{98.19} & \multicolumn{1}{c|}{98.56}    & \multicolumn{1}{c|}{98.28}     & \multicolumn{1}{c|}{97.49}         & \multicolumn{1}{c|}{97.17}              & \multicolumn{1}{c|}{96.52}          & 95.37                   \\ \hline
\end{tabular}}
\end{center}
\end{table}

Figure~\ref{benign_filters} illustrates the application of Instagram filters as unseen benign transforms on invisible watermarked images. In this study, Instagram filters such as Brooklyn, Clarendon, and Aden are used. These filters are designed to enhance facial images with unique aesthetic effects, each altering images in distinctive ways to cater to diverse visual preferences and styles. Figure~\ref{benign_filters} illustrates the application of multiple Instagram filters and their combined impact as unseen benign transforms on invisible watermarked images. In this study, Instagram filters such as Brooklyn, Clarendon, and Aden are used. The symbol (A+B) in the figure denotes the combined application of the Brooklyn and Aden filters to the watermarked image. Similarly, (B+C) in the figure denotes the combined application of Brooklyn and Clarendon filters to the watermarked image. Finally, (A+B+C) in the figure denotes the combined application of Aden, Brooklyn, and Clarendon filters to the watermarked image. The evaluation shows that the model incorporating U-Net with a Critic ($C$) and an Adversary ($A_{adv}$) network performs best in terms of Bit Recovery Accuracy (BRA). The overall performance in terms of BRA is $98.73\%$ for  U-Net+C+$A_{adv}$ on unseen benign transformations, which is better than $98.29\%$ BRA of the second-best model, Facesigns (Semi-Fragile), when only benign transformations $g_{bt}$ are used during the training stage. This superior performance is consistently observed across various settings that involve different image filters. 
The consistent performance of the model across all unknown benign transformations can be attributed to its specialized training exclusively on benign transformations such as cropping, compression, or subtle filtering. Training specifically on these types of transformations enables the model to become highly proficient in identifying and managing the specific patterns and distortions they introduce. %As a result, when the model is tested using similar benign transformations, it exhibits exceptional performance, effectively recognizing and adapting to these familiar alterations.

\textbf{Overall}, training the model using only benign transformations render it robust to unseen benign transformations. Further, the integration of the adversary module during the training stage also plays a pivotal role in enhancing the robustness and imperceptibility of the watermarking process to unknown benign transformations. This process renders the watermark resilient against a variety of benign transformations, thereby preserving the integrity of the media content.

\subsection{Malicious Transforms}
For malicious transformations, we applied facial manipulations on the watermarked images using different generative models, namely, auto-encoder, GANs, and diffusion models, and calculated the Bit recovery accuracy (BRA). In this case, a \textbf{low BRA} is preferred for fragility against malicious transformations such as Deepfakes.  \\
%Specifically, we used the Faceswap model which is an auto-encoder-based model that aligns the facial landmarks to swap the facial regions between the pair of images of the two identities\footnote{https://github.com/deepfakes/faceswap}.
%Additionally, we used FSGAN, StarGAN, and AttGAN as GAN-based facial identity swap and attribute manipulation techniques. We also used Stable Diffusion for generating diffusion-based facial manipulations to ensure that our approach incorporates the latest techniques in facial manipulation. %This use of cutting-edge technology helps maintain the relevance and effectiveness of our methods in handling contemporary digital media challenges.

\begin{figure*}[htbp]
\centerline{\includegraphics[width=0.60\textwidth]{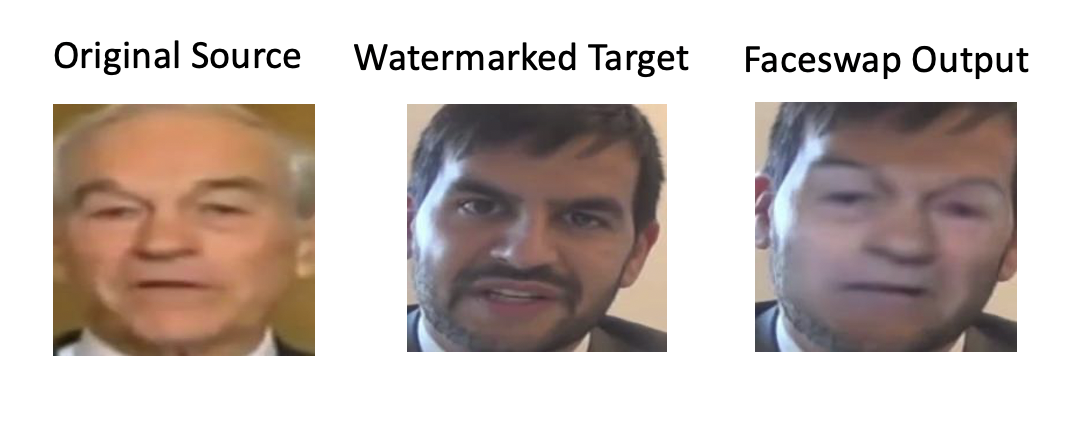}}
\caption{Example of sample watermarked facial images with identity swapped using the Faceswap model. The input to the Faceswap model is the source ($x_{sw}$) (not watermarked) and the target image ($x_{tw}$) (watermarked). The output is the malicious transformed facial image $x_{mt}=g_{mt}(x_{sw},x_{tw})$ with the identity swapped between the source and the target.
} 
\label{fig14}
\end{figure*}

\noindent \textbf{FaceSwap Based Malicious Transforms:}
The Faceswap model is a graphics-based method that aligns the facial landmarks to swap the faces between the source and the target using an encoder and decoder style model.  This technology is widely used for various applications ranging from entertainment and media to more serious uses such as personalized advertisements and synthetic data generation for AI training.
%In a face swap model, we have an encoder and decoder which play crucial roles in the overall process of swapping faces between two images. 
Figure~\ref{fig14} shows the sample watermarked facial images with identity swaps generated from the Faceswap model. The input to the Faceswap model is the source image ($x_{sw}$) (not watermarked) and target image ($x_{tw}$) (watermarked). The output is the maliciously transformed facial image $x_{mt}=g_{mt}(x_{sw},x_{tw})$ with the identity swapped between the source and the target. For detailed technical description and
implementation details, please refer to face-swap-based malicious transforms~\footnote{https://github.com/deepfakes/faceswap}.

\begin{table}[!htbp]

\caption{The effect of Faceswap encoder-decoder-based malicious transformation on the invisible watermarked target images, obtained using our proposed models in Table~\ref{TableConfig}, in terms of Bit recovery accuracy (BRA). All models are trained on the FF++ dataset.}
\label{Table10}
%\begin{center}
\begin{center}
\begin{tabular}{l|l|lc}
\hline
\multirow{2}{*}{Testing Dataset} & \multicolumn{1}{c|}{\multirow{2}{*}{Method}} & \multicolumn{2}{c}{BRA($\%$)}                             \\ \cline{3-4} 
                                 & \multicolumn{1}{c|}{}                        & \multicolumn{1}{l|}{None}  & \multicolumn{1}{l}{Faceswap} \\ \hline
\multirow{7}{*}{FF++}            & SemiFragile DCT                              & \multicolumn{1}{l|}{99.43} & 84.29                        \\
                                 & Hidden                                       & \multicolumn{1}{l|}{97.65} & 79.26                        \\
                                 & StegaStamp                                   & \multicolumn{1}{l|}{\textbf{99.62}} & 96.39                        \\
                                 & FaceSigns (Semi-Fragile)                     & \multicolumn{1}{l|}{99.49} & 43.84                        \\ \cline{2-4} 
                                 & U-Net+C+$A_{adv}$ (Baseline)                 & \multicolumn{1}{l|}{99.29} & 52.71                        \\
                                 & U-Net+C+$A_{adv}$ ($g_{bt}$)                 & \multicolumn{1}{l|}{99.45} & 63.39                        \\
                                 & U-Net+C+$A_{adv}$ ($g_{bt}$+$g_{mt}$)        & \multicolumn{1}{l|}{99.31} & \textbf{41.28}                       \\ \hline
\multirow{7}{*}{CelebA}         & SemiFragile DCT                              & \multicolumn{1}{c|}{99.28} & 85.45                        \\
                                 & Hidden                                       & \multicolumn{1}{c|}{97.46} & 71.64                        \\
                                 & StegaStamp                                   & \multicolumn{1}{c|}{\textbf{99.51}} & 95.56                        \\
                                 & FaceSigns (Semi-Fragile)                     & \multicolumn{1}{l|}{99.27} & \textbf{43.14}                        \\ \cline{2-4} 
                                 & U-Net+C+$A_{adv}$ (Baseline)                 & \multicolumn{1}{c|}{98.59} & 54.59                        \\
                                 & U-Net+C+$A_{adv}$ ($g_{bt}$)                 & \multicolumn{1}{c|}{99.16} & 66.72                        \\
                                 & U-Net+C+$A_{adv}$ ($g_{bt}$+$g_{mt}$)        & \multicolumn{1}{c|}{98.82} & 43.97                        \\ \hline
\end{tabular}
\end{center}
\end{table}

\begin{table}[!htbp]

\caption{The effect of Faceswap encoder-decoder-based malicious transformation on the invisible watermarked target images, obtained using our proposed models, in terms of Bit recovery accuracy (BRA). All models are trained on the CelebA dataset.}
\label{Table11}
%\begin{center}
\begin{center}
\begin{tabular}{l|l|lc}
\hline
\multirow{2}{*}{Testing Dataset} & \multicolumn{1}{c|}{\multirow{2}{*}{Method}} & \multicolumn{2}{c}{BRA($\%$)}                             \\ \cline{3-4} 
                                 & \multicolumn{1}{c|}{}                        & \multicolumn{1}{l|}{None}  & \multicolumn{1}{l}{Faceswap} \\ \hline
\multirow{7}{*}{FF++}            & SemiFragile DCT                              & \multicolumn{1}{l|}{99.29} & 81.53                        \\
                                 & Hidden                                       & \multicolumn{1}{l|}{97.73} & 78.24                        \\
                                 & StegaStamp                                   & \multicolumn{1}{l|}{\textbf{99.55}} & 94.72                        \\
                                 & FaceSigns (Semi-Fragile)                     & \multicolumn{1}{l|}{99.18} & 45.58                        \\ \cline{2-4} 
                                 & U-Net+C+$A_{adv}$ (Baseline)                 & \multicolumn{1}{l|}{99.36} & 52.79                        \\
                                 & U-Net+C+$A_{adv}$ ($g_{bt}$)                 & \multicolumn{1}{l|}{99.25} & 64.37                        \\
                                 & U-Net+C+$A_{adv}$ ($g_{bt}$+$g_{mt}$)        & \multicolumn{1}{l|}{98.69} & \textbf{42.72}                        \\ \hline
\multirow{7}{*}{CelebA}         & SemiFragile DCT                              & \multicolumn{1}{c|}{99.51} & 84.69                        \\
                                 & Hidden                                       & \multicolumn{1}{c|}{98.12} & 75.42                        \\
                                 & StegaStamp                                   & \multicolumn{1}{c|}{\textbf{99.67}} & 96.83                        \\
                                 & FaceSigns (Semi-Fragile)                     & \multicolumn{1}{l|}{99.38} & 38.32                        \\ \cline{2-4} 
                                 & U-Net+C+$A_{adv}$ (Baseline)                 & \multicolumn{1}{c|}{99.18} & 49.88                        \\
                                 & U-Net+C+$A_{adv}$ ($g_{bt}$)                 & \multicolumn{1}{c|}{99.48} & 62.11                        \\
                                 & U-Net+C+$A_{adv}$ ($g_{bt}$+$g_{mt}$)        & \multicolumn{1}{c|}{99.27} & \textbf{36.52}                        \\ \hline
\end{tabular}
\end{center}
\end{table}

Table~\ref{Table10} shows the effect of malicious transformations based on FaceSwap (based on the encoder-decoder model) on invisible watermarked target images, obtained using our proposed model, in terms of bit recovery accuracy (BRA). All models are trained on the FF++ dataset and tested on the FF++ and CelebA datasets. From the Table, overall performance in terms of (BRA) is $42.62\%$ for the U-Net+C+$A_{adv}$ on Faceswap-based malicious transforms which is lower than $43.49\%$ BRA of second best model FaceSigns (Semi-Fragile), when both benign $g_{bt}$ and malicious transformations $g_{mt}$ are used during training.

A lower BRA indicates an increased fragility, which is particularly valuable in the context of detecting malicious transformations like Deepfakes. These results are consistent for the FaceSwap model in both intra- and cross-dataset settings. The superior performance in terms of low BRA  can be attributed to the malicious transform used during the training stage. This shows that the transform is able to mimic the facial manipulation process where the facial features are perturbed.
Similarly, Table~\ref{Table11} shows the BRA for Faceswap model-based malicious transformations when trained on Celeb A and tested on FF++, and CelebA datasets. Again, overall performance in terms of (BRA) is $39.62\%$ for the U-Net+C+$A_{adv}$ on Faceswap-based malicious transforms which is lower than $41.95\%$ BRA of second best model FaceSigns (Semi-Fragile), when both benign $g_{bt}$ and malicious transformations $g_{mt}$ are used during training. These results are consistent for the Faceswap model in both intra- and cross-dataset settings. \\

\noindent \textbf{GAN-based Malicious Transforms:} For this experiment, we have used three popular GAN variants namely, FSGAN~\cite{Nirkin2019FSGANSA} (identity swap), StarGAN~\cite{Choi2017StarGANUG}, and AttGAN~\cite{He2017AttGANFA} (attribute manipualtion) for the generation of manipulated images. These GANs are widely used for identity, expression, and attribute-based facial manipulation generation. For detailed technical description and implementation details, please refer to the original paper on FSGAN~\cite{Nirkin2019FSGANSA}, StarGAN~\cite{Choi2017StarGANUG}, and AttGAN~\cite{He2017AttGANFA}. This information is not included for the sake of space.

%Table~\ref{Table5} shows the BRA of different watermarking techniques when GAN based malicious transformations are applied.

\begin{figure*}[htbp]
\centerline{\includegraphics[width=0.60\textwidth]{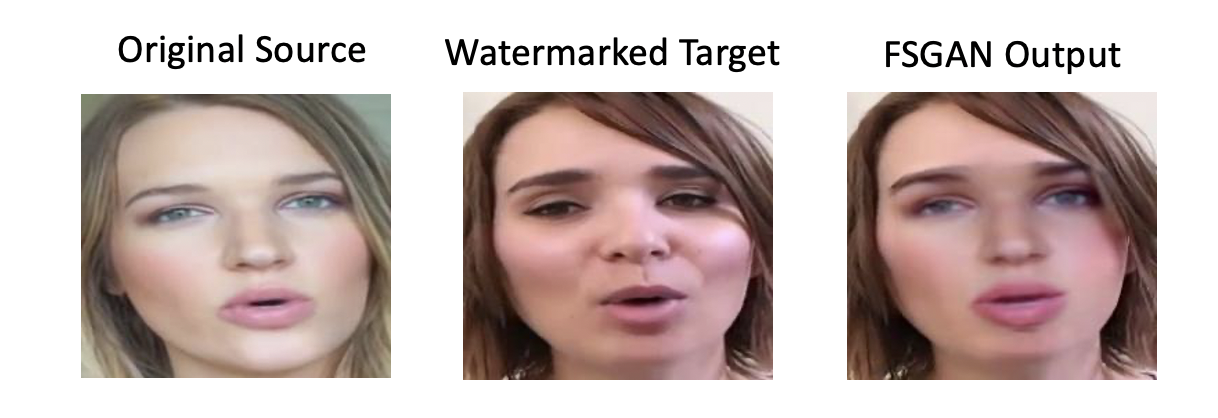}}
\caption{Example of sample watermarked facial images with identity swapping-based manipulation applied using the FSGAN model. The input to the FSGAN is the source image ($x_{sw}$) (not watermarked) and the target image ($x_{tw}$) (watermarked). The output is the maliciously transformed facial image $x_{mt}=g_{mt}(x_{w})$ with the identity swapped between the source and the target.
} 
\label{FSGAN}
\end{figure*}

\begin{figure}[htbp]
\centerline{\includegraphics[width=0.5\textwidth]{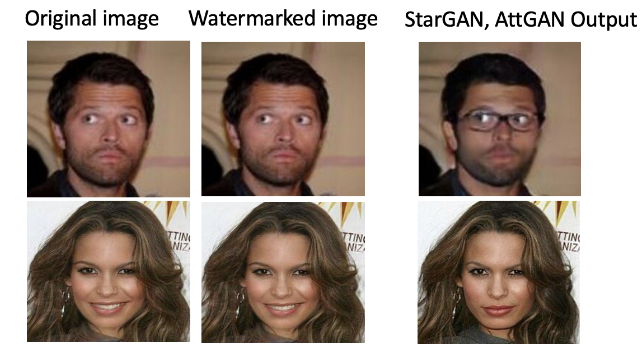}}
\caption{Example of sample watermarked facial images with attribute manipulations generated using the StarGAN and AttGAN models. The input to the StarGAN and AttGAN models is the watermarked image $x_w$ and the output is the manipulated watermarked facial image $x_{mt}=g_{mt}(x_{w})$ with the facial attributes, such as eyeglasses, facial expression, and hair color, edited.}
\label{fig12}
\end{figure}

Table~\ref{Table5} shows the effect of FSGAN~\cite{Nirkin2019FSGANSA}, StarGAN~\cite{Choi2017StarGANUG} and AttGAN~\cite{He2017AttGANFA} based facial manipulations, in terms of Bit recovery accuracy (BRA), on invisible watermarked facial images. These manipulations are applied to all three different versions of our model, as mentioned in Table~\ref{TableConfig}. Figure~\ref{FSGAN} gives an example of sample watermarked target facial images with identity swapped generated from the FSGAN model. The input to the FSGAN is the source image ($x_{sw}$) (not watermarked) and the target image ($x_{tw}$) (watermarked). The output is the malicious facial image transformed $x_{mt}=g_{mt}(x_{w})$ with the identity swapped between the source and the target. Similarly, Figure~\ref{fig12} gives the example of sample watermarked facial images with attribute manipulations generated from the StarGAN and AttGAN models. The input to the StarGAN and AttGAN models is the watermarked image $x_w$ and the output is the malicious transformed facial image $x_{mt}=g_{mt}(x_{w})$ with the manipulated facial attributes such as eye glasses, facial expression and hair color.

All these models, including our U-Net-based and GAN-based models, were trained on the FF++ dataset and then tested on both the FF++ and CelebA datasets. As can be seen from the Table, overall performance in terms of (BRA) is $49.01\%$ for the U-Net+C+$A_{adv}$ on GAN-based malicious transforms which is lower than $50.91\%$ BRA of second-best model FaceSigns (Semi-Fragile), when both benign $g_{bt}$ and malicious transformations $g_{mt}$ are used during training.
These results are consistent across FSGAN, StarGAN, and AttGAN models in the intra as well as cross-dataset settings.
Similarly, in Table~\ref{Table6} we used the same FSGAN, StarGAN, and AttGAN models for malicious transformations. The models are trained on CelebA and tested on FF++ and CelebA datasets. Again, the overall performance in terms of (BRA) is $47.39\%$ for the U-Net+C+$A_{adv}$ on GAN-based malicious transforms which is lower than $49.88\%$ BRA of the second-best model, FaceSigns (Semi-Fragile), when both benign $g_{bt}$ and malicious transformations $g_{mt}$ are used during training.

 The superior performance of the model, indicated by a low Bit Recovery Accuracy (BRA), can indeed be traced back to the inclusion of malicious transformations during the training phase. This approach allows the model to become adept at detecting and responding to manipulations akin to those encountered in real-world scenarios, such as Deepfakes and other forms of digital forgery. In addition, we trained the suggested model exclusively on malicious transformations; nonetheless, its performance is not as good as that of the model trained on both benign and malicious transformations. The reason could be that a model trained solely on malicious transformations tends to develop a narrow focus, optimizing specifically for certain types of data alterations. This specialization may limit the model's capacity to generalize across a wider array of real-world scenarios, which could include benign transformations. Lacking exposure to these benign transformations, the model may struggle to accurately differentiate between genuinely malicious modifications and benign transformations in images, resulting in decreased overall performance. \\

 %These results are consistent across FSGAN, StarGAN and AttGAN models in the intra as well as cross dataset settings. 

\begin{table}
\caption{The effect of FSGAN, StarGAN, and AttGAN-based facial manipulations on the invisible watermarked images in terms of Bit recovery accuracy (BRA). These malicious transformations are applied to watermarked images obtained using different variants of the proposed model as described in Table~\ref{TableConfig}. All the models are trained on the FF++ Dataset.}
\label{Table5}
\begin{center}
\scalebox{0.75}{
\begin{tabular}{l|l|lccc}
\hline
\multirow{2}{*}{Testing Dataset} & \multicolumn{1}{c|}{\multirow{2}{*}{Method}} & \multicolumn{4}{c}{Generative technique (BRA$\%$)}                                                                \\ \cline{3-6} 
                                 & \multicolumn{1}{c|}{}                        & \multicolumn{1}{l|}{None}  & \multicolumn{1}{l}{FSGAN} & \multicolumn{1}{l}{StarGAN} & \multicolumn{1}{l}{AttGAN} \\ \hline
\multirow{7}{*}{FF++}            & SemiFragile DCT                              & \multicolumn{1}{l|}{99.43} & 68.54                     & 59.78                       & 64.68                      \\
                                 & Hidden                                       & \multicolumn{1}{l|}{97.65} & 74.96                     & 67.45                       & 72.86                      \\
                                 & StegaStamp                                   & \multicolumn{1}{l|}{\textbf{99.62}} & 96.52                     & 97.41                       & 96.38                      \\
                                 & FaceSigns (Semi-Fragile)                     & \multicolumn{1}{l|}{99.49} & 51.49                     & 50.89                       & 48.14                      \\ \cline{2-6} 
                                 & U-Net+C+$A_{adv}$ (Baseline)                 & \multicolumn{1}{l|}{99.29} & 51.76                     & 50.14                       & 48.67                      \\
                                 & U-Net+C+$A_{adv}$ ($g_{bt}$)                 & \multicolumn{1}{l|}{99.45} & 65.35                     & 70.63                       & 61.29                      \\
                                 & U-Net+C+$A_{adv}$ ($g_{bt}$+$g_{mt}$)        & \multicolumn{1}{l|}{99.31} & \textbf{50.29}                     & \textbf{48.27}                       & \textbf{45.71}                      \\ \hline
\multirow{7}{*}{CelebA}         & SemiFragile DCT                              & \multicolumn{1}{c|}{99.28} & 66.86                     & 62.14                       & 65.41                      \\
                                 & Hidden                                       & \multicolumn{1}{c|}{97.46} & 76.29                     & 66.83                       & 74.19                      \\
                                 & StegaStamp                                   & \multicolumn{1}{c|}{\textbf{99.51}} & 97.23                     & 96.84                       & 97.52                      \\
                                 & FaceSigns (Semi-Fragile)                     & \multicolumn{1}{l|}{99.27} & 53.29                     & 52.28                       & 49.38                      \\ \cline{2-6} 
                                 & U-Net+C+$A_{adv}$ (Baseline)                 & \multicolumn{1}{c|}{98.59} & 53.43                     & 52.08                       & 50.73                      \\
                                 & U-Net+C+$A_{adv}$ ($g_{bt}$)                 & \multicolumn{1}{c|}{99.16} & 67.29                     & 72.73                       & 64.58                      \\
                                 & U-Net+C+$A_{adv}$ ($g_{bt}$+$g_{mt}$)        & \multicolumn{1}{c|}{98.82} & \textbf{51.83}                     & \textbf{50.69}                       & \textbf{47.28}                      \\ \hline
\end{tabular}
}
\end{center}
\end{table}

\begin{table}
\caption{The effect of FSGAN, StarGAN, and AttGAN-based facial manipulations on the invisible watermarked images in terms of Bit recovery accuracy (BRA). These malicious transformations are applied to watermarked target images obtained using different variants of the proposed model as described in Table~\ref{TableConfig}. All the models are trained on the CelebA Dataset.}
\label{Table6}
\begin{center}
\scalebox{0.73}{
\begin{tabular}{l|l|lccc}
\hline
\multirow{2}{*}{Testing Dataset} & \multicolumn{1}{c|}{\multirow{2}{*}{Method}} & \multicolumn{4}{c}{Generative technique (BRA$\%$)}                                                                \\ \cline{3-6} 
                                 & \multicolumn{1}{c|}{}                        & \multicolumn{1}{l|}{None}  & \multicolumn{1}{l}{FSGAN} & \multicolumn{1}{l}{StarGAN} & \multicolumn{1}{l}{AttGAN} \\ \hline
\multirow{7}{*}{FF++}            & SemiFragile DCT                              & \multicolumn{1}{l|}{99.29} & 67.84                     & 64.61                       & 65.74                      \\
                                 & Hidden                                       & \multicolumn{1}{l|}{97.73} & 76.41                     & 69.28                       & 74.17                      \\
                                 & StegaStamp                                   & \multicolumn{1}{l|}{\textbf{99.55}} & 97.72                     & 98.15                       & 97.28                      \\
                                 & FaceSigns (Semi-Fragile)                     & \multicolumn{1}{l|}{99.18} & 50.16                     & 52.78                       & 50.54                      \\ \cline{2-6} 
                                 & U-Net+C+$A_{adv}$ (Baseline)                 & \multicolumn{1}{l|}{98.69} & 52.57                     & 54.26                       & 52.25                      \\
                                 & U-Net+C+$A_{adv}$ ($g_{bt}$)                 & \multicolumn{1}{l|}{99.39} & 68.28                     & 73.86                       & 71.52                      \\
                                 & U-Net+C+$A_{adv}$ ($g_{bt}$+$g_{mt}$)        & \multicolumn{1}{l|}{99.17} & \textbf{48.53}                     & \textbf{51.29}                       & \textbf{47.17}                      \\ \hline
\multirow{7}{*}{CelebA}         & SemiFragile DCT                              & \multicolumn{1}{c|}{99.51} & 69.86                     & 62.64                       & 63.82                      \\
                                 & Hidden                                       & \multicolumn{1}{c|}{98.12} & 78.14                     & 68.18                       & 72.69                      \\
                                 & StegaStamp                                   & \multicolumn{1}{c|}{\textbf{99.67}} & 97.58                     & 97.64                       & 95.48                      \\
                                 & FaceSigns (Semi-Fragile)                     & \multicolumn{1}{l|}{99.38} & 47.27                     & 50.97                       & 47.56                      \\ \cline{2-6} 
                                 & U-Net+C+$A_{adv}$ (Baseline)                 & \multicolumn{1}{c|}{99.18} & 49.65                     & 52.54                       & 50.49                      \\
                                 & U-Net+C+$A_{adv}$ ($g_{bt}$)                 & \multicolumn{1}{c|}{99.48} & 67.53                     & 72.12                       & 67.28                      \\
                                 & U-Net+C+$A_{adv}$ ($g_{bt}$+$g_{mt}$)        & \multicolumn{1}{c|}{99.27} & \textbf{45.63}                     & \textbf{48.48}                       & \textbf{43.29}                     \\ \hline
\end{tabular}    
}
\end{center}
\end{table}

\begin{figure}[htbp]
\centerline{\includegraphics[width=0.50\textwidth]{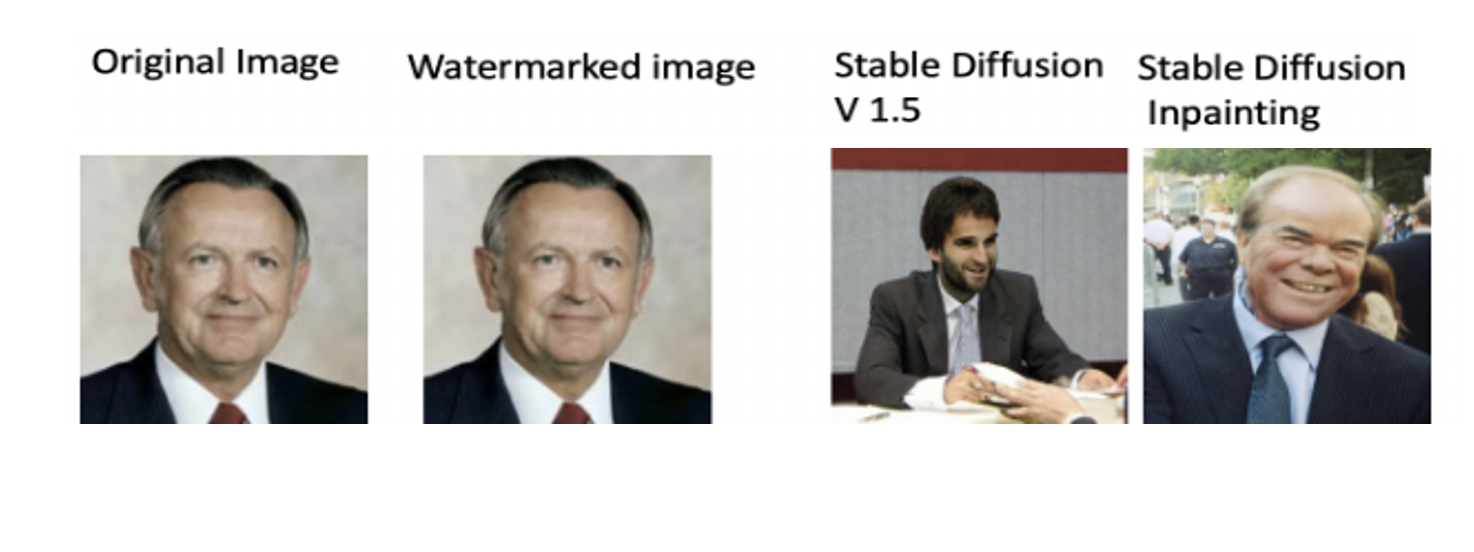}}
\caption{Example of Stable Diffusion V $1.5$ and inpainting models based malicious transformation on facial images. The input to the Stable Diffusion V $1.5$ and inpainting models is the watermarked image $x_w$ and the output is the maliciously transformed facial image $x_{mt}=g_{mt}(x_{w})$.}
\label{fig13}
\end{figure}

\noindent\textbf{Diffusion model based Malicious transforms:} In this work, we use a Stable diffusion-based model which is a latent text-to-image/image-to-image diffusion model able to take any type of text input and produce realistic-looking images. Figure~\ref{fig13} shows the example of maliciously transformed facial images from Stable Diffusion V $1.5$ and in-painting. The input to the Stable Diffusion V $1.5$ and in-painting models is the watermarked image $x_w$ and the output is a de-noised synthetic facial image $x_{mt}=g_{mt}(x_{w})$.

%\noindent \textbf{Stable Diffusion V 1.5}: In our experiments, Stable Diffusion V $1.5$ is used for image-to-image generation which is similar to text-to-image generation. But in this case, you have the option to provide a prompt as well as an initial image to initiate the diffusion process. The initial image undergoes encoding to be represented in latent space, followed by the addition of noise. Subsequently, the latent diffusion model, given both the prompt and the noisy latent image, predicts the added noise and then subtracts the predicted noise from the initial latent image to generate a new latent image. Finally, a decoder translates the new latent image back into an image.

%\noindent \textbf{Stable Diffusion Inpainting}: Inpainting involves the replacement or modification of specific regions within an image, making it a valuable tool for tasks such as image restoration, defect removal, or even the substitution of image areas with entirely new content. In this process, a mask is used to designate the regions of the image that require filling. The area designated for inpainting is typically indicated by white pixels in the mask, while the areas to be retained remain black. The white pixels are then filled in accordance with the provided prompt, allowing for the completion or alteration of the designated regions within the image.

\begin{table}
\caption{The effect of Stable diffusion V $1.5$ (SD 1.5) and Stable Diffusion based Inpainting models (SD Inpainting) for malicious transformations on the invisible watermarked images, in terms of Bit recovery accuracy (BRA), using different versions of our proposed models as given in Table~\ref{TableConfig}. All these models, including our U-Net-based models and Diffusion models, are trained on the IMDB-WIKI dataset and tested on the FF++ dataset.}
%on our proposed model (baseline), proposed model when only benign transformations $g_{bt}$ are used during training and proposed model when both benign $g_{bt}$ and malicious transformations $g_{mt}$ are used during training. All these models, including our U-Net based models and Diffusion models, were trained on IMDB-WIKI Dataset and tested on FF++ Dataset.}
\label{Table7}
\begin{center}
\scalebox{0.70}{
\begin{tabular}{c|l|c|cc}
\hline
\multicolumn{1}{l|}{\multirow{2}{*}{Testing Dataset}} & \multicolumn{1}{c|}{\multirow{2}{*}{Method}} & \multirow{2}{*}{Model}                              & \multicolumn{2}{l}{Generative technique (BRA$\%$)} \\ \cline{4-5} 
\multicolumn{1}{l|}{}                                 & \multicolumn{1}{c|}{}                        &                                                     & \multicolumn{1}{c|}{None}     & Stable Diffusion   \\ \hline
\multirow{14}{*}{FF++}                                & SemiFragile DCT                              & \multirow{7}{*}{SD 1.5}                             & \multicolumn{1}{c|}{98.78}    & 52.59              \\
                                                      & Hidden                                       &                                                     & \multicolumn{1}{c|}{97.56}    & 54.74              \\
                                                      & StegaStamp                                   &                                                     & \multicolumn{1}{c|}{\textbf{99.16}}    & 61.87              \\
                                                      & FaceSigns (Semi-Fragile)                     &                                                     & \multicolumn{1}{c|}{98.58}    & 49.81              \\ \cline{2-2} \cline{4-5} 
                                                      & U-Net+C+$A_{adv}$ (Baseline)                 &                                                     & \multicolumn{1}{c|}{99.07}    & 55.76              \\
                                                      & U-Net+C+$A_{adv}$ ($g_{bt}$)                 &                                                     & \multicolumn{1}{c|}{98.22}    & 60.76              \\
                                                      & U-Net+C+$A_{adv}$ ($g_{bt}$+$g_{mt}$)        &                                                     & \multicolumn{1}{c|}{98.31}    & \textbf{37.29}              \\ \cline{2-5} 
                                                      & SemiFragile DCT                              & \multicolumn{1}{l|}{\multirow{7}{*}{SD Inpainting}} & \multicolumn{1}{c|}{98.78}    & 58.87              \\
                                                      & Hidden                                       & \multicolumn{1}{l|}{}                               & \multicolumn{1}{c|}{97.56}    & 54.69              \\
                                                      & StegaStamp                                   & \multicolumn{1}{l|}{}                               & \multicolumn{1}{c|}{\textbf{99.16}}    & 69.26              \\
                                                      & FaceSigns (Semi-Fragile)                     & \multicolumn{1}{l|}{}                               & \multicolumn{1}{c|}{98.58}    & 55.67              \\ \cline{2-2} \cline{4-5} 
                                                      & U-Net+C+$A_{adv}$ (Baseline)                 & \multicolumn{1}{l|}{}                               & \multicolumn{1}{c|}{99.07}    & 59.52              \\
                                                      & U-Net+C+$A_{adv}$ ($g_{bt}$)                 & \multicolumn{1}{l|}{}                               & \multicolumn{1}{c|}{98.22}    & 67.85              \\
                                                      & U-Net+C+$A_{adv}$ ($g_{bt}$+$g_{mt}$)        & \multicolumn{1}{l|}{}                               & \multicolumn{1}{c|}{98.31}    & \textbf{46.72}              \\ \hline
\end{tabular}
}
\end{center}
\end{table}

\begin{table}
\caption{The effect of Stable diffusion V 1.5 and Stable Diffusion Inpainting based malicious transformations on the invisible watermarked images, obtained using different versions of our proposed models as given in Table~\ref{TableConfig}, in terms of Bit recovery accuracy (BRA). All the models are trained on the IMDB-WIKI Dataset and tested on the CelebA Dataset.}
\label{Table8}
\begin{center}
\scalebox{0.70}{
\begin{tabular}{c|l|c|cc}
\hline
\multicolumn{1}{l|}{\multirow{2}{*}{Testing Dataset}} & \multicolumn{1}{c|}{\multirow{2}{*}{Method}} & \multirow{2}{*}{Model}                              & \multicolumn{2}{l}{Generative technique (BRA$\%$)} \\ \cline{4-5} 
\multicolumn{1}{l|}{}                                 & \multicolumn{1}{c|}{}                        &                                                     & \multicolumn{1}{c|}{None}     & Stable Diffusion   \\ \hline
\multirow{14}{*}{CelebA}                             & SemiFragile DCT                              & \multirow{7}{*}{SD 1.5}                             & \multicolumn{1}{c|}{99.24}    & 54.29              \\
                                                      & Hidden                                       &                                                     & \multicolumn{1}{c|}{97.09}    & 57.24              \\
                                                      & StegaStamp                                   &                                                     & \multicolumn{1}{c|}{\textbf{99.52}}    & 70.17              \\
                                                      & FaceSigns (Semi-Fragile)                     &                                                     & \multicolumn{1}{c|}{99.11}    & 50.52              \\ \cline{2-2} \cline{4-5} 
                                                      & U-Net+C+$A_{adv}$ (Baseline)                 &                                                     & \multicolumn{1}{c|}{99.32}    & 54.59              \\
                                                      & U-Net+C+$A_{adv}$ ($g_{bt}$)                 &                                                     & \multicolumn{1}{c|}{98.92}    & 63.82              \\
                                                      & U-Net+C+$A_{adv}$ ($g_{bt}$+$g_{mt}$)        &                                                     & \multicolumn{1}{c|}{98.56}    & \textbf{38.74}              \\ \cline{2-5} 
                                                      & SemiFragile DCT                              & \multicolumn{1}{l|}{\multirow{7}{*}{SD Inpainting}} & \multicolumn{1}{c|}{99.24}    & 60.73              \\
                                                      & Hidden                                       & \multicolumn{1}{l|}{}                               & \multicolumn{1}{c|}{97.09}    & 58.26              \\
                                                      & StegaStamp                                   & \multicolumn{1}{l|}{}                               & \multicolumn{1}{c|}{\textbf{99.52}}    & 67.81              \\
                                                      & FaceSigns (Semi-Fragile)                     & \multicolumn{1}{l|}{}                               & \multicolumn{1}{c|}{99.11}    & 53.15              \\ \cline{2-2} \cline{4-5} 
                                                      & U-Net+C+$A_{adv}$ (Baseline)                 & \multicolumn{1}{l|}{}                               & \multicolumn{1}{c|}{99.32}    & 57.18              \\
                                                      & U-Net+C+$A_{adv}$ ($g_{bt}$)                 & \multicolumn{1}{l|}{}                               & \multicolumn{1}{c|}{98.92}    & 65.39              \\
                                                      & U-Net+C+$A_{adv}$ ($g_{bt}$+$g_{mt}$)        & \multicolumn{1}{l|}{}                               & \multicolumn{1}{c|}{98.56}    & \textbf{47.59}              \\ \hline
\end{tabular}}

\end{center}
\end{table}

\begin{table}
\caption{The effect of Stable diffusion V $1.5$ and Stable Diffusion Inpainting based malicious transformations on the invisible watermarked images, obtained using different versions of our proposed models as given in Table~\ref{TableConfig}, in terms of Bit recovery accuracy (BRA). All the models are trained and tested on the IMDB-WIKI Dataset.}
\label{Table9}
\begin{center}
\scalebox{0.70}{
\begin{tabular}{c|l|c|cc}
\hline
\multicolumn{1}{l|}{\multirow{2}{*}{Testing Dataset}} & \multicolumn{1}{c|}{\multirow{2}{*}{Method}} & \multirow{2}{*}{Model}                              & \multicolumn{2}{l}{Generative technique (BRA$\%$)} \\ \cline{4-5} 
\multicolumn{1}{l|}{}                                 & \multicolumn{1}{c|}{}                        &                                                     & \multicolumn{1}{c|}{None}     & Stable Diffusion   \\ \hline
\multirow{14}{*}{IMDB-WIKI}                           & SemiFragile DCT                              & \multirow{7}{*}{SD 1.5}                             & \multicolumn{1}{c|}{99.16}    & 54.57              \\
                                                      & Hidden                                       &                                                     & \multicolumn{1}{c|}{97.23}    & 52.84              \\
                                                      & StegaStamp                                   &                                                     & \multicolumn{1}{c|}{\textbf{99.41}}    & 68.25              \\
                                                      & FaceSigns (Semi-Fragile)                     &                                                     & \multicolumn{1}{c|}{98.91}    & 49.24              \\ \cline{2-2} \cline{4-5} 
                                                      & U-Net+C+$A_{adv}$ (Baseline)                 &                                                     & \multicolumn{1}{c|}{99.24}    & 52.19              \\
                                                      & U-Net+C+$A_{adv}$ ($g_{bt}$)                 &                                                     & \multicolumn{1}{c|}{99.08}    & 61.86              \\
                                                      & U-Net+C+$A_{adv}$ ($g_{bt}$+$g_{mt}$)        &                                                     & \multicolumn{1}{c|}{98.85}    & \textbf{36.61}              \\ \cline{2-5} 
                                                      & SemiFragile DCT                              & \multicolumn{1}{l|}{\multirow{7}{*}{SD Inpainting}} & \multicolumn{1}{c|}{99.16}    & 62.64              \\
                                                      & Hidden                                       & \multicolumn{1}{l|}{}                               & \multicolumn{1}{c|}{97.23}    & 56.45              \\
                                                      & StegaStamp                                   & \multicolumn{1}{l|}{}                               & \multicolumn{1}{c|}{\textbf{99.41}}    & 74.24              \\
                                                      & FaceSigns (Semi-Fragile)                     & \multicolumn{1}{l|}{}                               & \multicolumn{1}{c|}{98.91}    & 51.71              \\ \cline{2-2} \cline{4-5} 
                                                      & U-Net+C+$A_{adv}$ (Baseline)                 & \multicolumn{1}{l|}{}                               & \multicolumn{1}{c|}{99.24}    & 57.79              \\
                                                      & U-Net+C+$A_{adv}$ ($g_{bt}$)                 & \multicolumn{1}{l|}{}                               & \multicolumn{1}{c|}{99.08}    & 64.65              \\
                                                      & U-Net+C+$A_{adv}$ ($g_{bt}$+$g_{mt}$)        & \multicolumn{1}{l|}{}                               & \multicolumn{1}{c|}{98.85}    & \textbf{44.94}              \\ \hline
\end{tabular}}
\end{center}
\end{table}

In our experiments, we used stable diffusion V $1.5$~\cite{ho2020denoising} and stable diffusion inpainting~\cite{ho2020denoising} based diffusion models which use the underlying concept of conditional mechanism and generative modeling of latent representation following a reverse Markov process. %In general, conditional mechanisms in models and systems are techniques that adjust the behavior or output of a process based on specific conditions or inputs. These mechanisms are foundational in fields ranging from artificial intelligence and machine learning to control systems and software engineering. The idea is to make a system's actions contingent on certain criteria to enhance functionality, improve efficiency, or meet specified requirements.
In order to make the training process more efficient and faster, we used Low-Rank Adaptation or (LoRA)~\footnote{https://huggingface.co/blog/lora} which is a simple training method that drastically lowers the total number of trainable parameters of specific computationally complex layers. %Lora is applied to specific layers of a pre-trained model, typically those that are responsible for significant computation, such as the convolutional or fully connected layers in diffusion models. 
Instead of modifying the entire weight matrix of a layer, LoRA introduces two low-rank matrices $A$ and 
$B$. These matrices are much smaller in size compared to the original weight matrix $W$ of the layer. As a result, LoRA training is considerably quicker and more memory-efficient, and smaller model weights are generated that are simpler to share and store. For detailed technical description and implementation details, please refer to the original work~\cite{ho2020denoising}, not discussed for the sake of space.

Table~\ref{Table7}, Table~\ref{Table8}, and Table~\ref{Table9} shows the effect of stable diffusion V $1.5$~\cite{ho2020denoising} and stable diffusion inpainting~\cite{ho2020denoising} based malicious transformations on the invisible watermarked images in terms of Bit recovery accuracy (BRA). The impact of these synthetic manipulations is evaluated on different versions of our proposed model as tabulated in Table~\ref{TableConfig}. All the models are trained on IMDB-WIKI Dataset and tested on FF++, IMDB-WIKI, and CelebA Datasets. From the Table~\ref{Table7}, the overall performance in terms of (BRA) is $42\%$ for the U-Net+C+$A_{adv}$ on Diffusion-based malicious transforms which is lower than $52.74\%$ BRA of second best model FaceSigns (Semi-Fragile), when both benign $g_{bt}$ and malicious transformations $g_{mt}$ are used during training. These results are consistent across Stable diffusion V $1.5$ and Stable Diffusion Inpainting models in the intra as well as cross-dataset settings.

%In \textbf{summary}, for all the experiments conducted, the model trained exclusively on benign transformations proved effective on unseen benign transforms. %This outcome is due to the model’s training regimen, which was focused solely on benign transformations such as mild cropping, slight compression, or subtle filtering. 
%Training a model exclusively with these transformations allows it to become highly skilled at recognizing and managing the specific patterns and distortions they introduce.% This specialized training equips the model to effectively handle similar types of noise or alterations that it was exposed to during its training phase. As a result, the model exhibits excellent performance when encountering similar benign transformations.
\textbf{Overall}, when exposed to malicious transformations, the proposed model that is trained on both benign and malicious transforms exhibits a lower Bit Recovery Accuracy (BRA), which is actually desirable. This lower BRA indicates increased fragility, a characteristic that is vital for the detection of altered media generated using malicious transformations like Deepfakes.
  %This effectiveness stems from the model’s training regimen, which included exposure to both benign and malicious transformations. By training the model with a variety of transformations, it becomes highly skilled at recognizing and managing the specific patterns and distortions associated with both types. This specialized training equips the model to effectively handle similar types of noise or alterations that it encountered during its training phase. Consequently, the model exhibits excellent performance when faced with similar kinds of transformations.
The model's performance is attributed to its training with both benign and malicious transformations, enabling it to acquire a comprehensive understanding of such transforms overall. We also experimented by training the model solely on malicious transformations, but its performance was notably inferior compared to the model trained on both benign and malicious transformations.

\section{Adversarial Attacks and Threat Model}
\label{Robustness and threat analysis}
\subsection{Adversarial Attacks for Watermark Removal}

%In this series of experiments, we tested our model against various adversarial attacks, including both white-box and black-box attacks. 

%While our model experienced a slight reduction in BRA post-adversarial attacks, we anticipate that the performance of traditional watermarking schemes would suffer more significantly under the same conditions.

To further understand the robustness and analyze potential threats against our proposed watermarking technique, we conducted evaluations of our model against specific adversarial attacks aimed at watermark removal. We did not reevaluate other existing watermarking methods under these adversarial conditions since our model had already demonstrated superior performance in terms of Bit Recovery Accuracy (BRA) under normal conditions over existing watermarking techniques. Further, a study in~\cite{zhao2023generative} documents the vulnerability of existing invisible watermarking techniques to watermark removal attacks.

Our evaluations have focused on both white-box and black-box scenarios, which are detailed as follows.\\

%For robustness and threat analysis of our proposed watermarking technique, we evaluated our proposed model on white-box and black-box-based watermark removal-based adversarial attacks explained as follows.

%The adversarial attacks are techniques used to fool machine learning models through malicious input~\cite{athalye2018obfuscated,zhao2023generative}. These attacks craft special inputs that are almost indistinguishable from natural data but cause the model to make mistakes. Adversarial attacks are primarily studied in the context of deep learning and have significant implications for the security of automated systems that rely on these technologies.

\noindent \textbf{White-box attacks}: In white-box attacks, the adversary has complete knowledge of the model, including its architecture and parameters~\cite{goodfellow2014explaining,carlini2017towards}. 
This access allows the attacker to precisely calculate the most effective perturbations to the input data to confuse the model. As the attacker has full information about the model, white-box attacks are generally considered more powerful and effective compared to black-box attacks. In this series of experiments involving adversarial attacks, we employ Bit Recovery Accuracy (BRA) and Detection Accuracy (DA) as evaluation metrics. %BRA is a measure of how accurately a system can recover a specific bit string from a manipulated media, while DA evaluates the system’s ability to accurately detect the presence of a watermark in the manipulated media. Therefore, both BRA and DA are crucial for assessing the robustness of a system when subjected to adversarial attacks, demonstrating its effectiveness in maintaining integrity and detecting alterations.
The popular white-box attacks used in this study are gradient-based methods namely the Fast Gradient Sign Method (FGSM)~\cite{goodfellow2014explaining}, Carlini \& Wagner Attack (C\&W)~\cite{carlini2017towards}, Backward Pass Differentiable Approximation (BPDA) and Expectation Over Transformation (EOT)~\cite{athalye2018obfuscated} that iteratively perturb input features to maximize the model's prediction error. These attacks are applied to the watermarked facial images to evaluate the robustness of our model against white-box attacks in terms of bit recovery accuracy (BRA) and detection accuracy (DA). For detailed technical description and implementation details on white-box attacks, please refer to the original work~\cite{goodfellow2014explaining,carlini2017towards,athalye2018obfuscated}.

%Here we use "bit recovery accuracy (BRA) " and "detection accuracy (DA) " as the performance metrics that are essential for evaluating the effectiveness of a system. In general Bit recovery accuracy (BRA) focuses on the precision of data extraction, which is critical in data-centric applications. Detection accuracy (DA), however, concerns the correct identification of the presence or absence of data, which is broader and applicable in initial detection phases.

%The Fast Gradient Sign Method (FGSM)~\cite{goodfellow2014explaining} utilizes the gradients of a neural network to generate an adversarial example. Similarly, the Carlini \& Wagner~\cite{carlini2017towards} Attack formulates an optimization problem to discover a perturbation that minimally alters the input while maximizing the model's prediction error. Advanced techniques such as BPDA (Backward Pass Differentiable Approximation) and EOT (Expectation Over Transformation)~\cite{athalye2018obfuscated} are employed in adversarial machine learning to create effective adversarial examples, especially in situations where direct gradient calculation is hindered by non-differentiable operations or transformations that introduce randomness.
\begin{figure}[htbp]
\centerline{\includegraphics[width=0.5\textwidth]{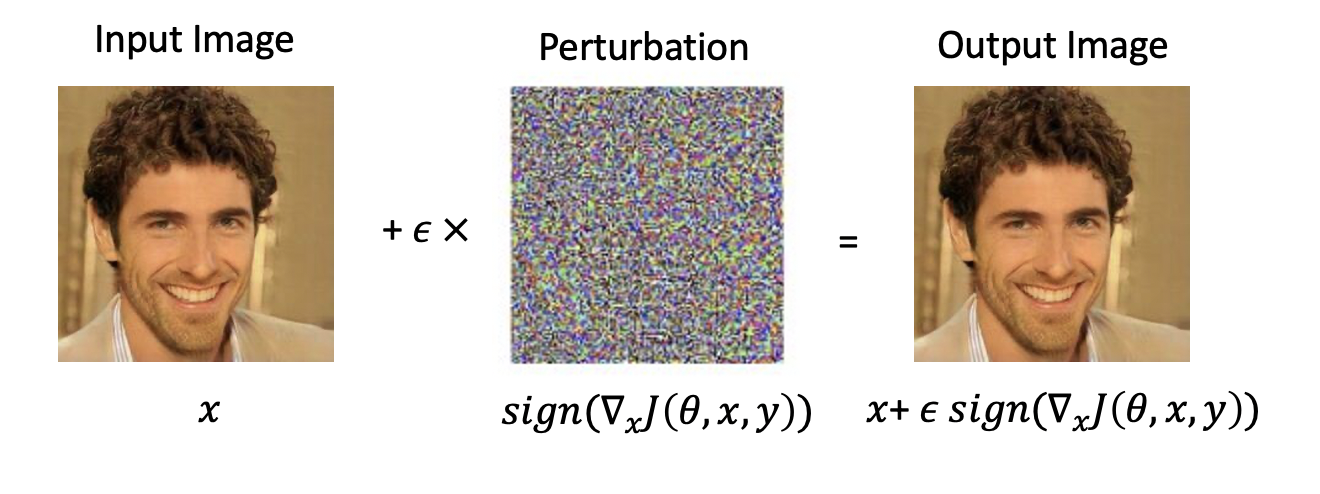}}
\caption{Example of an adversarial white-box attack using FGSM. Given an input image $x$, FGSM method utilizes the gradients of the loss function with respect to the input image to generate
a new image $x_{adv}$ that maximizes the prediction error.}
\label{fig15}
\end{figure}

%In our work, we used different types of white-box attacks namely, Fast gradient sign method , Carlini and Wagner Attack (C\&W), BPDA (Backward Pass Differentiable Approximation) and EOT (Expectation Over Transformation).

\begin{figure*}[htbp]
\centerline{\includegraphics[width=0.85\textwidth]{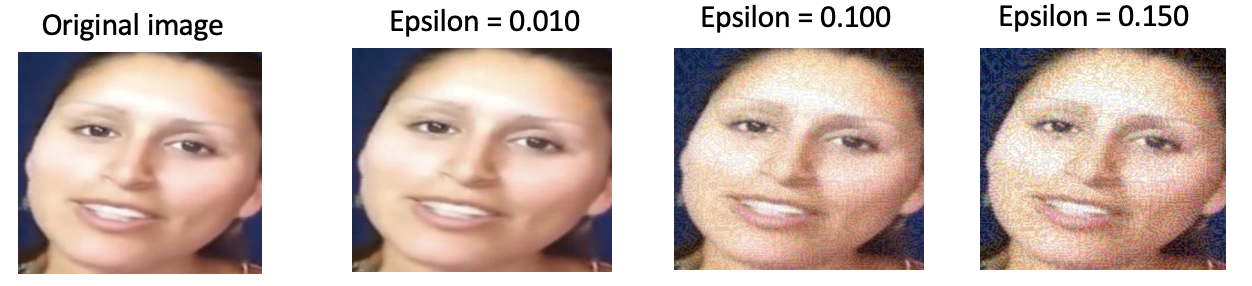}}
\caption{Example of sampled watermarked images generated using the FGSM-based white-box adversarial attack using different $\epsilon$. $\epsilon$ is the multiplier to control the magnitude of perturbation. All the experiments are based on $\epsilon$ value of $0.010$ to ensure perturbation imperceptibility [best viewed in Zoom].
} 
\label{fig16}
\end{figure*}

Figure~\ref{fig15} shows the example of an adversarial attack with FGSM, given an input image $x$, the FGSM method utilizes the gradients of the loss function of the individual classifiers from the Table~\ref{TableConfig} with respect to the input image to generate a new image $x_{adv}$ that maximizes the loss function. Similarly, Figure~\ref{fig16} shows the samples of watermarked images generated from the FGSM-based adversarial attack. Here $\epsilon$ is the multiplier to ensure the perturbations are small. In our experiments, we tested various values of $\epsilon$ to determine the effectiveness of the attack. We ultimately selected an $\epsilon$ value of $0.010$ for all our experiments involving the FGSM-based white-box attack. This particular value was chosen because it introduces distortions that are not visible to the human eye, ensuring that the modifications remain imperceptible while still assessing the system's robustness against adversarial attacks.
%A notable aspect of the Fast Gradient Sign Method (FGSM) is its utilization of gradients with respect to the input image. This choice stems from the objective of maximizing the loss when crafting an adversarial image. The strategy involves determining the contribution of each pixel in the image to the loss value and introducing a perturbation accordingly. This approach is computationally efficient because it allows for the straightforward calculation of how each input pixel influences the loss, facilitated by the chain rule and gradient computation. Consequently, gradients are computed with respect to the input image.

%Moreover, since the model is not undergoing training during this process—resulting in gradients not being computed with respect to the trainable variables, i.e., the model parameters—the model parameters remain unchanged. The sole aim is to deceive an already trained model, without modifying its parameters.

\begin{table}
\caption{The effect of FGSM, Carlini\&Wagner, BPDA\&EOT based white-box adversarial attacks on the invisible watermarked images in terms of Bit Recovery Accuracy~(BRA) on our proposed models. All the models are trained and tested on CelebA dataset.}
\label{Table12}
\begin{center}
\scalebox{0.8}{
\begin{tabular}{c|c|l|l}
\hline
\multicolumn{1}{l|}{Testing Dataset} & \multicolumn{1}{l|}{Adversarial Attack} & \multicolumn{1}{c|}{Method} & BRA(\%) \\ \hline
\multirow{12}{*}{Celeb A}            & \multirow{3}{*}{None}                   & U-Net                       & \textbf{99.57}   \\
                                     &                                         & U-Net+C                     & 99.43   \\
                                     &                                         & U-Net+C+$A_{adv}$           & 99.18   \\ \cline{2-4} 
                                     & \multirow{3}{*}{FGSM}                   & U-Net                       & 70.45   \\
                                     &                                         & U-Net+C                     & 71.68   \\
                                     &                                         & U-Net+C+$A_{adv}$           & \textbf{74.69}   \\ \cline{2-4} 
                                     & \multirow{3}{*}{Carlini\&Wagner}        & U-Net                       & 65.29   \\
                                     &                                         & U-Net+C                     & 66.53   \\
                                     &                                         & U-Net+C+$A_{adv}$           & \textbf{68.26}   \\ \cline{2-4} 
                                     & \multirow{3}{*}{BPDA\&EOT}           & U-Net                       & 56.62   \\
                                     &                                         & U-Net+C                     & 58.79   \\
                                     &                                         & U-Net+C+$A_{adv}$           & \textbf{62.64}   \\ \hline
\end{tabular}}
\end{center}
\end{table}

In our experiments, we combined both BPDA and EOT to render the attack very powerful. As BPDA can navigate through non-differentiable operations, EOT can ensure the adversarial example remains effective across a range of expected transformations~\cite{athalye2018obfuscated}. This combination is especially useful in attacking systems where input preprocessing and dynamic transformations are common, such as in vision-based machine learning models used in real-world scenarios.

Table~\ref{Table12} shows the effect of FGSM, carlini\&wagner, BPDA\&EOT based white-box based adversarial attacks on the invisible watermarked images generated using our proposed models (All the models are trained and tested on CelebA dataset) in terms of Bit recovery accuracy (BRA). As can be seen from the Table, the U-Net+C+$A_{adv}$ (Baseline) model outperforms the U-Net and U-Net+C baselines in terms of overall BRA performance. From the results, it is evident that the models, particularly U-Net+C+$A_{adv}$, maintain high BRA even under the challenging conditions posed by these white-box adversarial attacks. This enhanced performance of the U-Net+C+$A_{adv}$ model can be attributed to its integrated approach, which combines the basic capabilities of U-Net with the advanced refinement processes offered by the Critic (C) and Adversary ($A_{adv}$) modules. These additional components help improve the model's resilience against attacks by effectively learning to counteract the specific manipulations introduced by adversarial techniques, thus ensuring more robust watermark recovery.\\

%The underlying reason of resilience is the use of adversarial training which aims to remove/manipulate the watermark, while the encoder strives to preserve it. Through this adversarial interaction, the system is trained to embed robust watermarks that are difficult to remove or tamper with.\\ %This adversarial dynamic helps ensure that the watermark is not only present but also sufficiently robust to resist various types of adversarial attacks.

\noindent \textbf{Black-box attacks}: For black-box attacks, the attacker has limited or no access to the target model's details, such as its architecture or parameters~\cite{zhao2023generative}. Instead, the attacker can only interact with the model by providing inputs and observing the corresponding outputs.
%Black-box attacks represent a form of adversarial challenge in which the attacker has only limited insight into the target system's internal mechanisms, such as the specific algorithms, parameters, or the details of its training data. 
%In this scenario, the attacker only has access to the system's inputs and outputs. This access allows them to analyze how the system responds to various inputs and use this information to infer potential vulnerabilities or methods to undermine the system. %Such attacks are particularly common in environments where systems are accessible via an API, facilitating the attacker's ability to interact with the system and collect data.

%Attackers often leverage these interactions to build their datasets by repeatedly querying the system. Through these queries, they can accumulate enough response data to construct surrogate models. 

\begin{table}
\caption{The effect of VAE embedding and reconstruction-based black-box adversarial attack on the invisible watermarked images in terms of Bit recovery accuracy (BRA) of our proposed models. All the models are trained and tested on the CelebA dataset.}
\label{Table13}
\begin{center}
\scalebox{0.75}{
\begin{tabular}{c|c|l|l}
\hline
\multicolumn{1}{l|}{Testing Dataset} & \multicolumn{1}{l|}{Adversarial Attack (Black Box)} & \multicolumn{1}{c|}{Method} & BRA(\%) \\ \hline
\multirow{6}{*}{Celeb A}             & \multirow{3}{*}{None}                               & U-Net                       & \textbf{99.57}   \\
                                     &                                                     & U-Net+C                     & 99.43   \\
                                     &                                                     & U-Net+C+$A_{adv}$           & 99.18   \\ \cline{2-4} 
                                     & \multirow{3}{*}{VAE Embedding}                      & U-Net                       & 71.25   \\
                                     &                                                     & U-Net+C                     & 70.89   \\
                                     &                                                     & U-Net+C+$A_{adv}$           & \textbf{72.84}   \\ \hline
\end{tabular}}
\end{center}
\end{table}

\begin{figure}[htbp]
\centerline{\includegraphics[width=0.65\textwidth]{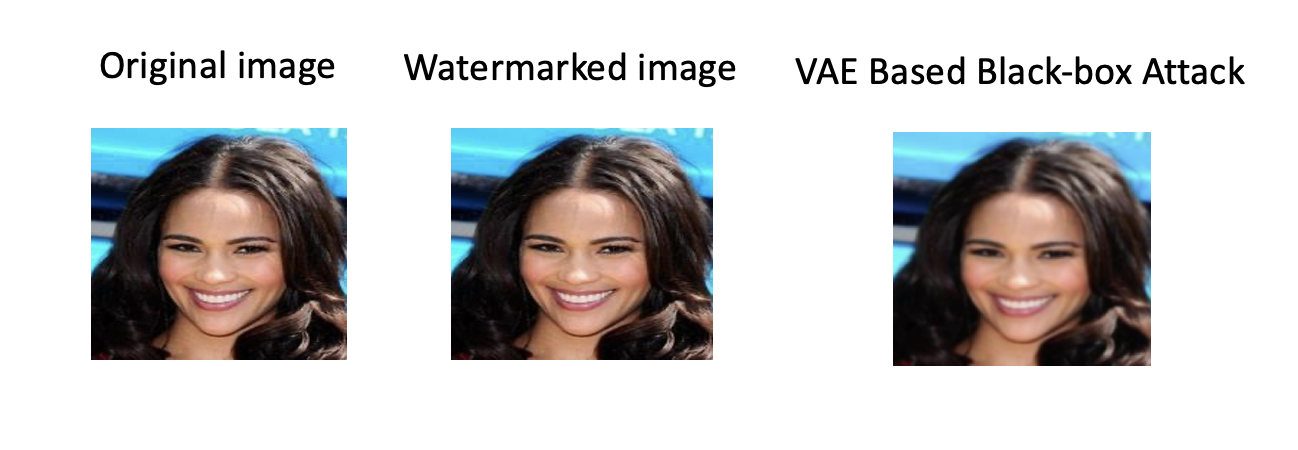}}
\caption{Example of sampled watermarked images generated after the VAE Embedding and Reconstruction-based black-box adversarial attack for watermark removal has been applied. The VAE-based attack is successful in removing the watermark to an extent however the attack over-smoothen's the image, resulting in blurriness [best viewed in Zoom].}
\label{fig17}
\end{figure}

In this study, we used the latest regeneration attacks proposed in~\cite{zhao2023generative} which aim to remove invisible watermarks. These attacks work by first adding random noise to the image to disrupt the watermark and then using image reconstruction techniques to restore the image quality. The authors instantiated the regeneration attacks into three instances namely, Identity Embedding with Denoising Reconstruction, VAE Embedding and Reconstruction, and Diffusion Embedding and Reconstruction. 
In our work, we employ VAE (Variational Autoencoder) based embedding and reconstruction for black-box attacks proposed in~\cite{zhao2023generative} because it is computationally less expensive and more flexible and efficient. In VAE Embedding and Reconstruction, we have VAEs which are trained using two different losses: a prior matching loss that constrains the latent to follow a pre-specified prior distribution, and a reconstruction loss that calculates the distance between the reconstructed and the original sample. For detailed technical description and implementation details, please refer to the original work~\cite{zhao2023generative}.

Figure~\ref{fig17} shows the sample of watermarked images after the application of the VAE Embedding and Reconstruction-based black-box watermark removal attack. The VAE-based attack is successful in removing the watermark only to an extent and at the same time the attack over-smooths the image, resulting in blurriness.

Table~\ref{Table13} shows the VAE embedding and reconstruction-based black-box adversarial attack on the invisible watermarked images in terms of Bit recovery accuracy (BRA) on our proposed models. All the models are trained and tested on the CelebA dataset. As can be seen from the Table, the U-Net+C+$A_{adv}$ (Baseline) model outperforms the U-Net and U-Net+C baselines in terms of overall BRA performance. The results clearly demonstrate that the U-Net+C+$A_{adv}$ model excels in maintaining high Bit Recovery Accuracy (BRA) even when faced with the rigorous demands of white-box adversarial attacks. The superior performance of this model can be traced back to its integrated design, which merges the foundational attributes of U-Net with the sophisticated enhancement capabilities provided by the Critic (C) and Adversary ($A_{adv}$) modules. These additional features enhance the model's robustness by enabling it to effectively respond to and neutralize the specific types of manipulations typical of adversarial attacks. Consequently, this ensures a more resilient process for watermark recovery, preserving the integrity of the watermarked images under adversarial conditions.

Additionally, while the Bit Recovery Accuracy (BRA) is lower than it was without adversarial attacks, it remains above the threshold necessary for detecting authentic media. This indicates that despite the impact of the attacks, the system's ability to verify authenticity through watermark recovery is still effective.

%So, finally 
\textbf{Overall}, the effectiveness of our proposed model to adversarial attacks again stems from the implementation of adversarial training, in which the adversary network attempts to remove the watermark, while the encoder strives to preserve it. This adversarial interaction trains the model to embed watermarks that are more difficult to remove or manipulate. Thus, this dynamic architecture not only guarantees the presence of the watermark but also significantly boosts its robustness, making it capable of withstanding a variety of adversarial attacks. This robustness is crucial for maintaining the integrity and security of the embedded data across different scenarios.

\subsection{Threat Model}
\label{Threat Model}

%\noindent \textbf{Threat Models:} 
Adversarial threats from attackers trying to avoid detection of altered media will be very likely encountered by our watermark embedding model. Next, we enlist a few potential threat scenarios that our model might face, and discuss possible solutions.\\

\noindent \textbf{Attack 1. Requesting information from the decoder network to launch hostile attacks:} Using an image, the attacker can query the decoder network to obtain the decoded message. Once the decoded message matches the target message, the attacker can manipulate the query image in an adversarial manner.

\noindent \textbf{Defense:} The Attacker lacks knowledge of the specific target messages that validate media authenticity, as these messages may be kept confidential and regularly updated. Even if the attacker obtains access to the secret message by querying the decoder with a watermarked image, the secrecy of the encryption key can prevent the attacker from identifying the target encrypted message for the decoder. Additionally, the decoder network can be securely hosted and is only capable of producing a binary label indicating whether the image is authentic or manipulated by comparing the decoded secret with a list of trusted secrets. Consequently, the signal from the decoder becomes impractical for executing adversarial attacks to match a target message from the vast pool of $2^{64}$ possible messages.

\noindent \textbf{Attack 2. A proxy encoder's training:} The attacker can collect an original and watermarked image dataset and use it to train an encoder-decoder neural network that performs image-to-image translation. Any new image can be successfully mapped by this network to a watermarked image.

\noindent \textbf{Defense:} To keep an attacker from obtaining a pair of original and watermarked images, one protection strategy is to store only watermarked images on devices. Furthermore, in order to enable the adversary to learn a generator for watermarking new images with the same secret message, the attack outlined above can only be executed if all of the encoded images have the same secret message. In order to combat this, some message components can be kept dynamic. These can include device-specific codes and a distinct timestamp, ensuring that every embedded bit-string is distinct. Another defense against such attacks is to update the encryption key or trustworthy message on a regular basis.

\noindent \textbf{Attack 3. Transferring the watermark perturbations between different images:} To verify the altered media, the adversary can try to extract the added perturbations of the watermark and apply them to a Deepfake image.

\noindent \textbf{Defense:} We speculate that, as our model produces a perturbation that is specific to a particular message, the decoder should not be able to retrieve the same perturbation when it is applied to other images. Through an experiment, we extract added perturbations from 50 watermarked images and apply the extracted perturbation to 50 alternate images in order to prove this notion. Such an attack has a bit recovery accuracy of only $18.5\%$, which is less accurate than random prediction.

\noindent \textbf{Attack 4. Model Inversion Attacks:} Attackers use output data from watermarking or Deepfake detection systems to reconstruct the original input data or sensitive attributes about the data, compromising privacy.

\noindent \textbf{Defense:} Incorporating differential privacy techniques during the training of watermarking and detection models helps safeguard the confidentiality of the input data by preventing the models from disclosing sensitive information. Additionally, introducing noise to the generated outputs of these systems further enhances privacy protection by ensuring that the output cannot be used to accurately reconstruct the input data, or reveal precise details about it.

\noindent \textbf{Attack 5. Side-Channel Attacks:} An attacker exploits side-channel information such as computation time, power consumption, or electromagnetic emissions to gain insights into the watermarking or detection algorithms, potentially revealing secret keys or operations.

\noindent \textbf{Defense:} To mitigate timing attacks, it is essential to design algorithms that execute in constant time, ensuring that their operation duration does not vary based on the input. This approach prevents attackers from deducing sensitive information based on how long the algorithm takes to process different inputs. Additionally, to safeguard against side-channel attacks, implementing physical security measures such as shielding techniques and restricting physical access to systems is crucial.

\noindent \textbf{Attack 8. Reverse Engineering Attacks:} Attackers deconstruct the watermarking or Deepfake detection system to understand its mechanism fully. With this knowledge, they could develop more effective methods to remove or bypass watermarks or to create more convincing Deepfakes that evade detection.

\noindent \textbf{Defense:} Applying code obfuscation techniques to make reverse engineering more difficult and time-consuming. Utilizing secure hardware environments like Trusted Execution Environments (TEE) to run critical parts of the watermarking or detection processes, shielding them from reverse engineering attempts.

\section{Ablation study}
\label{Ablation}
In this section, an ablation study is conducted to assess the individual contributions of various modules within the proposed model, which consists of different configurations of the U-Net architecture enhanced with additional modules like Critic and Adversary. This methodical approach allows for a clearer understanding of how each component affects the overall performance of the model. 

Note that in this series of experiments, we utilized the baseline model along with the critic and adversarial modules, which were not trained on either benign or malicious transformations. The purpose of employing only the baseline model without exposing it to these transformations was to clearly demonstrate the fundamental capabilities and limitations of each module prior to any influence from benign or malicious changes. This is crucial for analyzing the inherent effectiveness of each component within our proposed model, providing a foundational understanding of their impact before considering the additional complexities introduced by specific transformations. This kind of analysis is essential for systems where understanding the discrete contribution of each component is key to overall performance and reliability.

\begin{table}
\caption{Ablation study on the impact of each module (network) used in our proposed model. }
\label{Table14}
\begin{center}
\scalebox{1.0}{
\begin{tabular}{c|l|cc}
\hline
\multirow{2}{*}{Testing Dataset} & \multicolumn{1}{c|}{\multirow{2}{*}{Method}} & \multicolumn{2}{c}{BRA(\%)}           \\ \cline{3-4} 
                                 & \multicolumn{1}{c|}{}                        & \multicolumn{1}{c|}{None}  & Faceswap \\ \hline
\multirow{3}{*}{Celeb A}         & U-Net                                        & \multicolumn{1}{c|}{99.57} & 54.68    \\
                                 & U-Net+C                                      & \multicolumn{1}{c|}{99.43} & 51.82    \\
                                 & U-Net+C+$A_{adv}$                            & \multicolumn{1}{c|}{99.18} & 49.88    \\ \hline
\end{tabular}}
\end{center}
\end{table}

 \begin{enumerate}
\item \noindent \textbf{U-Net:} Initially, the U-Net model along with the discriminator is trained without benign and malicious transformations and without the integration of any Critic or Adversary modules. %No transformations are applied during training. 
This setup serves as the control group, providing a benchmark to measure the impact of adding the other modules.
\item \noindent \textbf{U-Net+C:} To this baseline, a Critic module is added, creating a second variant of the model. The Critic module is designed to assess the quality of the output and guide the network towards generating more realistic images.
\item \noindent \textbf{U-Net+C+$A_{adv}$:} The most complex variant includes both the Critic and the Adversary modules alongside the baseline U-Net. The Adversary module simulates potential attacks or challenges the model might face, aiming to ensure that the watermarks are robust against various types of manipulations, particularly those that might be encountered in adversarial environments.
\end{enumerate}

The results, as shown in~\ref{Table14}, provide a comparative analysis of the impact of each configuration when tested on the CelebA dataset. Specifically, the study finds that the model equipped with both the Critic and Adversary modules (U-Net+C+$A_{adv}$) shows superior performance in Bit Recovery Accuracy (BRA). This configuration notably excels when evaluated on malicious transformation.% typically used in Faceswap encoder-decoder scenarios on watermarked images.

The introduction of the Critic and Adversary modules to the model influences the Bit Recovery Accuracy (BRA) in various scenarios. The Critic module, aimed at enhancing visual fidelity, can compromise the (BRA), as it may prioritize image quality over the exactness of watermark retrieval. This effect is noted both under normal conditions and for malicious transformation i.e., Faceswap, where the Critic helps maintain realistic reconstructions but may lower BRA by prioritizing visual authenticity. Similarly, the Adversary module, which simulates attacks to test robustness, can lower BRA by making the watermark more secure but harder to decode in standard conditions. This module proves particularly useful in strengthening the system’s resilience against malicious changes, such as those in Faceswap, yet this robustness can also complicate watermark extraction, leading to lower BRA in typical detection scenarios. The addition of benign and malicious transformations during the training stage will add enhance the robustness and fragility of our model to benign and malicious transformations, respectively, as already discussed in the previous set of experiments.

Based on the enhanced performance observed, the U-Net model augmented with both Critic and Adversary modules is selected as the baseline model for all subsequent experiments. This decision is based on the model’s demonstrated ability to handle both benign and malicious transformations effectively, ensuring higher fidelity in watermark recovery under adversarial conditions. Thus, this methodical approach not only emphasizes the importance of each module but also demonstrates how integrating these modules can result in substantial enhancements in the robustness and accuracy of our proposed model for image watermarking. This is especially pertinent in situations where maintaining the integrity and authenticity of digital content is paramount.

\section{Conclusion and Future work}
\label{conclusion}
With the volume of Deepfakes showing staggering growth, advanced proactive defense mechanisms are required for media authentication and to control misinformation spread in advance.
In this paper, we introduce a novel deep learning-based semi-fragile invisible image watermarking technique as a proactive defense that allows media authentication by verifying an invisible secret message embedded in the image pixels.
%can certify the integrity of digital 
%images and reliably detect facial manipulations based Deepfakes. 
Our proposed approach systematically integrates a U-Net-based encoder-decoder style architecture with the discriminator, critic, and adversarial network for efficient watermark embedding and robustness against watermark removal. 
Thorough experimental investigations on popular facial Deepfake datasets demonstrate that our proposed watermarking technique generates highly imperceptible watermarks that are recoverable with high bit recovery accuracy under benign image processing operations. Further, the watermark is not recoverable when facial manipulations based Deepfakes, generated using different generative algorithms, are applied. Cross comparison with the existing invisible image watermarking techniques proves the efficacy of our proposed approach in terms of imperceptibility and bit recovery accuracy.
In addition, the watermarked images obtained using our proposed model are resilient to several white-box and black-box watermark removal attacks. This is attributed to the adversarial network used during the training stage that mimics the efforts of an adversary in removing the watermark embedded by the encoder, thereby obtaining resilience. Thus advancing the state-of-the-art in watermarking as a proactive defense for media authentication and for combating Deepfakes.
%Through our experiments and evaluations, we demonstrate that 
%our proposed approach generates more imperceptible watermarks than previous state-of-the-art methods while upholding
%the desired semi-fragile characteristics and resilience to watermark removal. 
%By carefully designing a fixed set of differentiable benign and malicious
%transformations during training, our framework achieves generalizability to real-world image transformations
%and can reliably detect Deepfake facial manipulations, unlike existing image watermarking techniques.
Our proposed watermarking technique can be vital to media authenticators in social media platforms, news agencies, and legal offices and help create more trustworthy and responsible platforms and establish consumer trust in digital media. Our work has two primary limitations. Firstly, the complexity of our model necessitates advanced hardware and GPU support, which we plan to address in future iterations by optimizing the model for improved generalizability. Secondly, we were unable to simulate all potential attacks outlined in the threat model discussed in the section~\ref{Threat Model}. As a part of future work, we aim to address these limitations. Further, we plan to extend our proposed semi-fragile technique for watermarking multi-modal audio-visual data streams in videos.

\small{
\flushend
\bibliographystyle{ACM-Reference-Format}
\bibliography{Main}}

\end{document}